\documentclass[final]{cvpr}

\usepackage{times}
\usepackage{epsfig}
\usepackage{graphicx}
\usepackage{amsmath}
\usepackage{amssymb}
\usepackage{mathtools}
\usepackage{booktabs}
\usepackage{xcolor}
\usepackage{subcaption}
\usepackage{bm}

\usepackage[pagebackref=true,breaklinks=true,colorlinks,bookmarks=false]{hyperref}

\newcommand{\modelname}{{SceneGen}}
\newcommand{\ourdataset}{{ATG4D}}

\begin{document}

\title{SceneGen: Learning to Generate Realistic Traffic Scenes}

\author{
Shuhan Tan$^{1,2}$\thanks{Indicates equal contribution. Work done at Uber ATG.}
\quad Kelvin Wong$^{1,3*}$ \quad Shenlong Wang$^{1,3}$ \\
Sivabalan Manivasagam$^{1,3}$ \quad Mengye Ren$^{1,3}$ \quad Raquel Urtasun$^{1,3}$ \\
$^{1}$Uber Advanced Technologies Group \quad $^{2}$Sun Yat-Sen University \quad $^{3}$University of Toronto \\
{
    \tt\small{tanshh@mail2.sysu.edu.cn} \quad
    \tt\small{\{kelvinwong,slwang,manivasagam,mren,urtasun\}@cs.toronto.edu}
}
}

\maketitle


\begin{abstract}
We consider the problem of generating realistic traffic scenes automatically.
Existing methods typically insert actors into the scene according to a set
of hand-crafted heuristics and are limited in their ability to model the true
complexity and diversity of real traffic scenes, thus inducing a
content gap between synthesized traffic scenes versus real ones.
As a result, existing simulators lack the fidelity necessary
to train and test self-driving vehicles.
To address this limitation, we present SceneGen---a neural autoregressive model
of traffic scenes that eschews the need for rules and heuristics.
In particular, given the ego-vehicle state and a high definition map of surrounding
area, {\modelname} inserts actors of various classes into
the scene and synthesizes their sizes, orientations, and velocities.
We demonstrate on two large-scale datasets
{\modelname}'s ability to faithfully model distributions of
real traffic scenes.
Moreover, we show that {\modelname} coupled with sensor simulation
can be used to train perception models that generalize to the real world.
\end{abstract}


\section{Introduction}
The ability to simulate realistic traffic scenarios is an important milestone
on the path towards safe and scalable self-driving.
It enables us to build rich virtual environments in which we can improve our
self-driving vehicles (SDVs) and verify their safety and performance~\cite{dosovitskiy2017,sumo2018,manivasagam2020,wong2020}.
This goal, however, is challenging to achieve.
As a first step, most large-scale self-driving programs simulate pre-recorded
scenarios captured in the real world~\cite{manivasagam2020} or employ teams of test engineers
to design new scenarios~\cite{dosovitskiy2017,sumo2018}.
Although this approach can yield realistic simulations, it is ultimately not
scalable.
This motivates the search for a way to generate realistic traffic scenarios \emph{automatically}.

More concretely, we are interested in generating the layout of actors
in a traffic scene given the SDV's current state and a high
definition map (HD map) of the surrounding area.
We call this task \emph{traffic scene generation} (see Fig.~\ref{figure:teaser}).
Here, each actor is parameterized by a class label, a bird's eye view bounding box,
and a velocity vector.
Our lightweight scene parameterization is popular among existing self-driving simulation{
stacks and can be readily used in downstream modules; \eg, to simulate LiDAR~\cite{dosovitskiy2017,fang2019,manivasagam2020}.

A popular approach to traffic scene generation is to use procedural models to
insert actors into the scene according to a set of rules~\cite{yang1996,sumo2018,dosovitskiy2017,prakash2019}.
These rules encode reasonable heuristics such as ``pedestrians should stay
on the sidewalk'' or ``vehicles should drive along lane centerlines'', and
their parameters can be manually tuned to give reasonable results.
Still, these simplistic heuristics cannot fully capture the complexity
and diversity of real world traffic scenes, thus inducing a content gap between
synthesized traffic scenes and real ones~\cite{kar2019}.
Moreover, this approach requires significant time and
expertise to design good heuristics and tune their
parameters.

\begin{figure}[t]
\includegraphics[width=\linewidth]{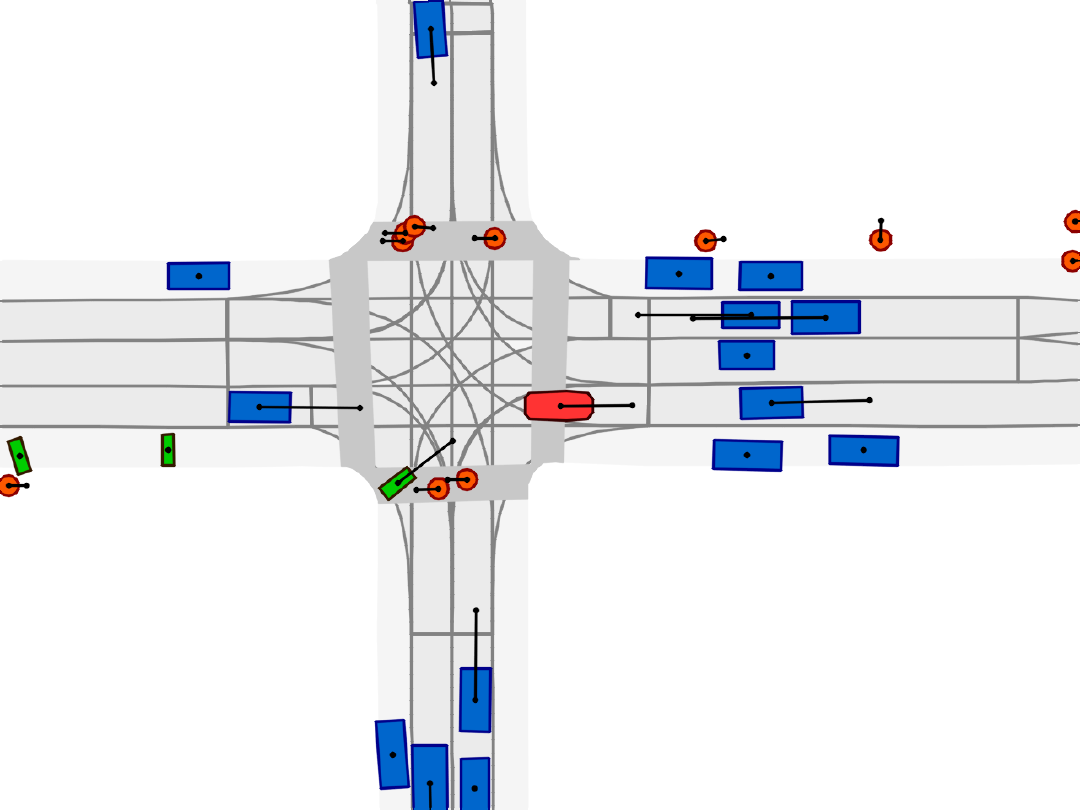}
\caption{
Given the SDV's state and an HD map, SceneGen autoregressively
inserts actors onto the map to compose a realistic traffic scene.
The ego SDV is shown in \textcolor{red}{red};
vehicles in \textcolor{blue}{blue}; pedestrians in \textcolor{orange}{orange};
and bicyclists in \textcolor{green}{green}.
}
\label{figure:teaser}
\end{figure}

To address these issues, recent methods use machine learning techniques to
automatically tune model parameters~\cite{wheeler2015,wheeler2016,jesenski2019,kar2019,devaranjan2020}.
These methods improve the realism and scalability of traffic scene generation.
However, they remain limited by their underlying hand-crafted heuristics and priors;
\eg, pre-defined scene grammars or assumptions about road topologies.
As a result, they lack the capacity to model the true complexity
and diversity of real traffic scenes and, by extension, the fidelity
necessary to train and test SDVs in simulation.
Alternatively, we can use a simple data-driven approach by sampling from
map-specific empirical distributions~\cite{fang2019}.
But this cannot generalize to new maps and may yield scene-inconsistent samples.

In this paper, we propose {\modelname}---a traffic scene generation model that
eschews the need for hand-crafted rules and heuristics.
Our approach is inspired by recent successes in deep generative modeling that
have shown remarkable results in estimating distributions of a variety of data,
without requiring complex rules and heuristics;
\eg, handwriting~\cite{graves2013}, images~\cite{oord2016a}, text~\cite{radford2019}, \etc.
Specifically, {\modelname} is a neural autoregressive model that, given the SDV's
current state and an HD map of the surrounding area, sequentially inserts actors
into the scene---mimicking the process by which humans do this as well.
As a result, we can sample realistic traffic scenes from SceneGen and
compute the likelihood of existing ones as well.

We evaluate {\modelname} on two large-scale self-driving datasets.
The results show that {\modelname} can better estimate the distribution over real
traffic scenes than competing baselines and generate more realistic samples as well.
Furthermore, we show that {\modelname} coupled with sensor simulation can generate
realistic labeled data to train perception models that generalize to the real world.
With {\modelname}, we take an important step towards developing SDVs
safely and scalably through large-scale simulation.
We hope our work here inspires more research along this direction so that one
day this goal will become a reality.


\section{Related Work}

\paragraph{Traffic simulation:}
The study of traffic simulation can be traced back to at least the 1950s
with Gerlough's dissertation on simulating freeway traffic flow~\cite{gerlough1955}.
Since then, various traffic models have been used for simulation.
Macroscopic models simulate entire populations of vehicles in the
aggregate~\cite{lighthill1955,richards1956} to study ``macroscopic'' properties
of traffic flow, such as traffic density and average velocity.
In contrast, microscopic models simulate the behavior of each individual vehicle
over time by assuming a car-following model~\cite{pipes1953,chandler1958,newell1961,gazis1961,gipps1976,bando1995,treiber2001}.
These models improve simulation fidelity considerably but at the cost of computational efficiency.
Microscopic traffic models have been included in popular software packages such as
SUMO~\cite{sumo2018}, CORSIM~\cite{owen2001}, VISSIM~\cite{fellendorf1994}, and MITSIM~\cite{yang1996}.

Recently, traffic simulation has found new applications in testing and training the autonomy stack of SDVs.
However, existing simulators do not satisfy
the level of realism necessary to properly test SDVs~\cite{wheeler2015}. For
example, the CARLA simulator~\cite{dosovitskiy2017} spawns actors at pre-determined locations and
uses a lane-following controller to simulate the vehicle behaviors over time. This approach is
too simplistic and so it induces a sim2real content gap~\cite{kar2019}. Therefore, in this paper,
we study how to generate snapshots of traffic scenes that mimic the realism and diversity of real ones.

\begin{figure*}
\includegraphics[width=\textwidth]{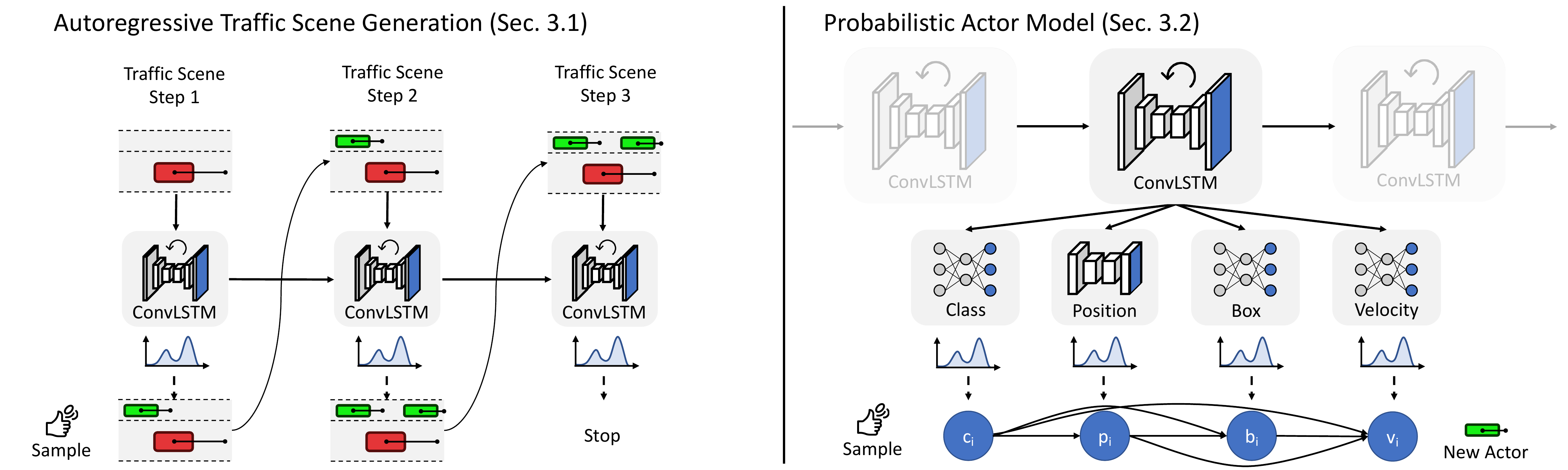}
\caption{
Overview of our approach.
Given the ego SDV's state and an HD map of the surrounding area, {\modelname}
generates a traffic scene by inserting actors one at a time (Sec.~\ref{section:method/autoregressive-factorization}).
We model each actor $ \bm{a}_i \in \mathcal{A} $ probabilistically, as a product over distributions of
its class $ c_i \in \mathbb{C} $, position $ \bm{p}_i \in \mathbb{R}^2 $, bounding
box $ \bm{b}_i \in \mathbb{B} $, and velocity $ \bm{v}_i \in \mathbb{R}^2 $ (Sec.~\ref{section:method/actor-model}).
}
\label{figure:concept}
\end{figure*}

\paragraph{Traffic scene generation:}
While much of the research into microscopic traffic simulation have focused
on modeling actors' behaviors, an equally important yet underexplored
problem is how to generate realistic snapshots of traffic scenes.
These snapshots have many applications; \eg, to initialize
traffic simulations~\cite{wheeler2015} or to generate labeled data for
training perception models~\cite{kar2019}.
A popular approach is to procedurally insert actors into the scene
according to a set of rules~\cite{yang1996,sumo2018,dosovitskiy2017,prakash2019}.
These rules encode reasonable heuristics such as ``pedestrians should stay on
the sidewalk'' and ``vehicles should drive along lane centerlines'', and their
parameters can be manually tuned to give reasonable results.
For example, SUMO~\cite{sumo2018} inserts vehicles into
lanes based on minimum headway requirements and initializes their speeds
according to a Gaussian distribution~\cite{wheeler2015}.
Unfortunately, it is difficult to scale this approach to new environments
since tuning these heuristics require significant time and expertise.

An alternative approach is to learn a probabilistic distribution over traffic scenes
from which we can sample new scenes~\cite{wheeler2015,wheeler2016,jesenski2019,fang2019,geiger2011a,geiger2011b,zhang2013}.
For example, Wheeler~\etal~\cite{wheeler2015} propose a Bayesian
network to model a joint distribution over traffic scenes in straight
multi-lane highways.
This approach was extended to model inter-lane dependencies~\cite{wheeler2016}
and generalized to handle a four-way intersection~\cite{jesenski2019}.
These models are trained to mimic a real distribution over traffic scenes.
However, they consider a limited set of road topologies only and assume
that actors follow reference paths in the map.
As a result, they are difficult to generalize to real urban scenes,
where road topologies and actor behaviors are considerably more complex;
\eg, pedestrians do not follow reference paths in general.

Recent advances in deep learning have enabled a more flexible approach to learn
a distribution over traffic scenes.
In particular, MetaSim~\cite{kar2019} augments the probabilistic scene graph
of Prakash~\etal \cite{prakash2019} with a graph neural network.
By modifying the scene graph's node attributes, MetaSim reduces the content gap
between synthesized images versus real ones, without manual tuning.
MetaSim2~\cite{devaranjan2020} extends this idea by learning to
sample the scene graph as well.
Unfortunately, these approaches are still limited by their hand-crafted scene
grammar which, for example, constrains vehicles to lane centerlines.
We aim to develop a more general method that avoids requiring these heuristics.

\paragraph{Autoregressive models:}
Autoregressive models factorize a joint distribution over $ n $-dimensions
into a product of conditional distributions $ p(x) = \prod_{i = 1}^{n} p(x_i | x_{< i}) $.
Each conditional distribution is then approximated with a parameterized
function~\cite{frey1995,bengio1999,uria2013a,uria2013b,uria2014}.
Recently, neural autoregressive models have found tremendous success in
modeling a variety of data, including handwriting~\cite{graves2013}, images~\cite{oord2016a},
audio~\cite{oord2016b}, text~\cite{radford2019}, sketches~\cite{ha2018},
graphs~\cite{liao2019}, 3D meshes~\cite{nash2020}, indoor scenes~\cite{wang2018} and
image scene layouts~\cite{jyothi2019}.
These models are particularly popular since they can factorize
a complex joint distribution into a product of much simpler conditional distributions.
Moreover, they generally admit a tractable likelihood, which can be used
for likelihood-based training, uncovering interesting/outlier examples, \etc.
Inspired by these advances, we exploit autoregressive models for
traffic scene generation as well.


\section{Traffic Scene Generation}

\begin{figure*}
\includegraphics[width=\textwidth]{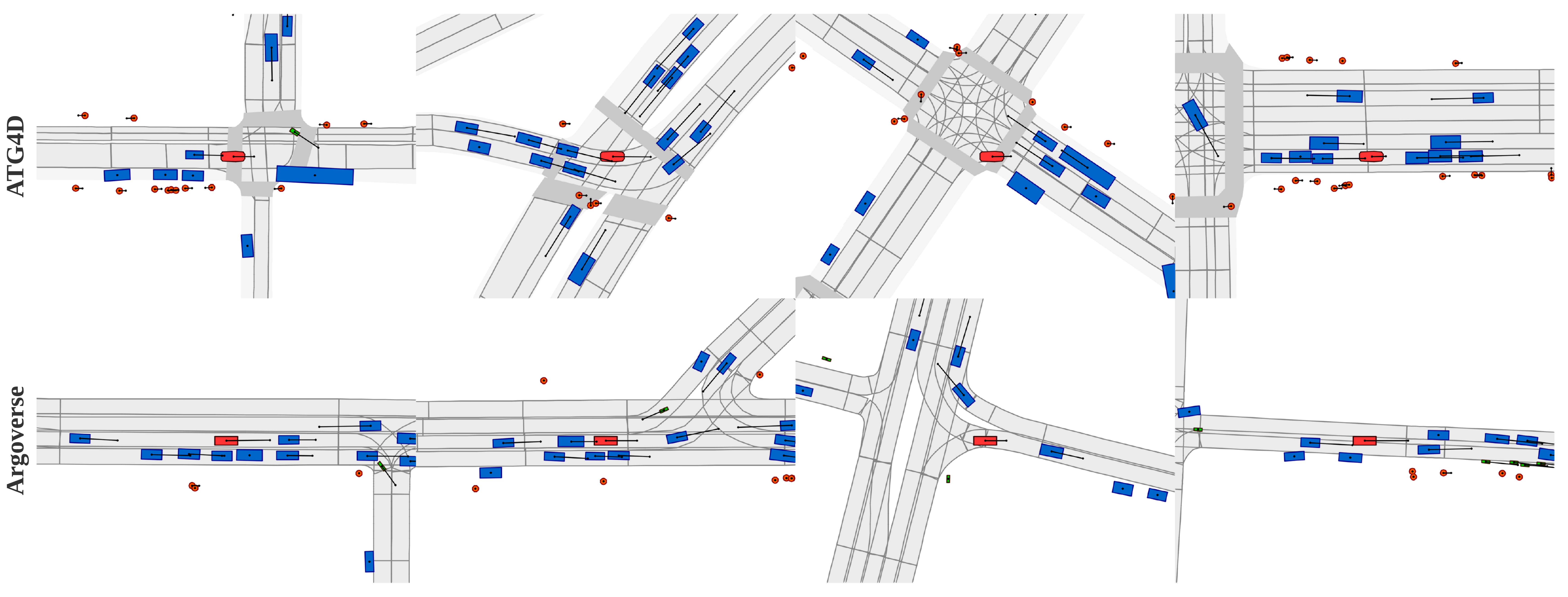}
\caption{
Traffic scenes generated by {\modelname} conditioned on HD maps from
{\ourdataset} (top) and Argoverse (bottom).
}
\label{figure:qualitative-results}
\end{figure*}

Our goal is to learn a distribution over traffic scenes from which we can
sample new examples and evaluate the likelihood of existing ones.
In particular, given the SDV $ \bm{a}_0 \in \mathcal{A} $ and an HD map
$ \bm{m} \in \mathcal{M} $, we aim to estimate the joint distribution over
other actors in the scene $ \{\bm{a}_1, \ldots, \bm{a}_n\} \subset \mathcal{A} $,
\begin{align}\label{equation:conditional-distribution}
    p(\bm{a}_1, \ldots, \bm{a}_n | \bm{m}, \bm{a}_0)
\end{align}

The HD map $ \bm{m} \in \mathcal{M} $ is a collection of
polygons and polylines that provide semantic priors for a
region of interest around the SDV;
\eg, lane boundaries, drivable areas, traffic light states.
These priors provide important contextual information about the scene and
allow us to generate actors that are consistent with the underlying road topology.

We parameterize each actor $ \bm{a}_i \in \mathcal{A} $ with an eight-dimensional random variable
containing its class label $ c_i \in \mathbb{C} $,
its bird's eye view location $ (x_i, y_i) \in \mathbb{R}^2 $,
its bounding box $ \bm{b}_i \in \mathbb{B} $\footnote{Pedestrians are not
represented by bounding boxes. They are represented by a single point
indicating their center of gravity.}, and its
velocity $ \bm{v}_i \in \mathbb{R}^2 $.
Each bounding box $ \bm{b}_i \in \mathbb{B} $ is a 3-tuple consisting of
the bounding box's size $ (w_i, l_i) \in \mathbb{R}^2_{> 0} $
and heading angle $ \theta_i \in [0, 2\pi) $.
In our experiments, $ \mathbb{C} $ consists of three classes:
vehicles, pedestrians, and bicyclists.
See Fig.~\ref{figure:teaser} for an example.

Modeling Eq.~\ref{equation:conditional-distribution} is a challenging task
since the actors in a given scene are highly correlated among themselves and
with the map, and the number of actors in the scene is random as well.
We aim to model Eq.~\ref{equation:conditional-distribution} such that our
model is easy to sample from and the resulting samples reflect the complexity
and diversity of real traffic scenes.
Our approach is to autoregressively factorize Eq.~\ref{equation:conditional-distribution}
into a product of conditional distributions.
This yields a natural generation process that sequentially inserts actors
into the scene one at a time.
See Fig.~\ref{figure:concept} for an overview of our approach.

In the following, we  first describe our autoregressive factorization
of Eq.~\ref{equation:conditional-distribution} and how we model this with a
recurrent neural network (Sec.~\ref{section:method/autoregressive-factorization}).
Then, in Sec.~\ref{section:method/actor-model}, we describe how {\modelname} generates
a new actor at each step of the generation process.
Finally, in Sec.~\ref{section:method/training},
we discuss how we train and sample from {\modelname}}.

\subsection{The Autoregressive Generation Process}
\label{section:method/autoregressive-factorization}

Given the SDV $ \bm{a}_0 \in \mathcal{A} $ and an HD map $ \bm{m} \in \mathcal{M} $,
our goal is to estimate a conditional distribution over the actors in the scene
$ \{\bm{a}_1, \ldots, \bm{a}_n\} \subset \mathcal{A} $.
As we alluded to earlier, modeling this conditional distribution is challenging
since the actors in a given scene are highly correlated among themselves and
with the map, and the number of actors in the scene is random.
Inspired by the recent successes of neural autoregressive models~\cite{graves2013,oord2016a,radford2019},
we propose to autoregressively factorize $ p(\bm{a}_1, \ldots, \bm{a}_n | \bm{m}, \bm{a}_0) $ into a
product of simpler conditional distributions.
This factorization simplifies the task of modeling the complex joint
distribution $ p(\bm{a}_1, \ldots, \bm{a}_n | \bm{m}, \bm{a}_0) $ and results in a model
with a tractable likelihood.
Moreover, it yields a natural generation process that mimics how a
human might perform this task as well.

In order to perform this factorization, we assume a fixed canonical ordering
over the sequence of actors $ \bm{a}_1, \ldots, \bm{a}_n $,
\begin{align}
    p(\bm{a}_1, \ldots, \bm{a}_n | \bm{m}, \bm{a}_0) = p(\bm{a}_{1} | \bm{m}, \bm{a}_0) \prod_{i = 1}^{n} p(\bm{a}_{i} | \bm{a}_{< i}, \bm{m}, \bm{a}_0)
\end{align}
where $ \bm{a}_{< i} = \{\bm{a}_{1}, \ldots, \bm{a}_{i - 1}\} $ is
the set of actors up to and including the $ i - 1 $-th actor in canonical order.
In our experiments, we choose a left-to-right, top-to-bottom
order based on each actor's position in bird's eye view coordinates.
We found that this intuitive ordering works well in practice.

Since the number of actors per scene is random, we introduce a stopping
token $ \bot $ to indicate the end of our sequential generation process.
In practice, we treat $ \bot $ as an auxillary actor that, when generated,
ends the generation process.
Therefore, for simplicity of notation, we assume that the last actor
$ \bm{a}_n $ is always the stopping token $ \bot $.

\paragraph{Model architecture:}
Our model uses a recurrent neural network to capture the
long-range dependencies across our autoregressive generation process.
The basis of our model is the ConvLSTM architecture~\cite{shi2015}---an
extension of the classic LSTM architecture~\cite{hochreiter1997} to spatial
data---and the input to our model at the $ i $-th generation step is a bird's
eye view multi-channel image encoding the SDV $ \bm{a}_0 $, the HD map $ \bm{m} $,
and the actors generated so far $ \{\bm{a}_{1}, \ldots, \bm{a}_{i - 1}\} $.

For the $ i $-th step of the generation process:
Let $ \bm{x}^{(i)} \in \mathbb{R}^{C \times H \times W} $ denote the multi-channel image,
where $ C $ is the number of feature channels and $ H \times W $ is the size of
the image grid.
Given the previous hidden and cell states $ \bm{h}^{(i - 1)} $ and $ \bm{c}^{(i - 1)} $,
the new hidden and cell states are given by:
\begin{align}
    \bm{h}^{(i)}, \bm{c}^{(i)} &= \mathrm{ConvLSTM}(\bm{x}^{(i)}, \bm{h}^{(i - 1)}, \bm{c}^{(i - 1)}; \bm{w}) \\
    \bm{f}^{(i)} &= \mathrm{CNN}_{\mathrm{b}}(\bm{h}^{(i)}; \bm{w})
\end{align}
where $ \mathrm{ConvLSTM} $ is a two-layer ConvLSTM,
$ \mathrm{CNN}_{\mathrm{b}} $ is a five-layer convolutional neural network (CNN) that extract backbone features,
and $ \bm{w} $ are the neural network parameters.
The features $ \bm{f}^{(i)} $ summarize the generated scene so far
$ \bm{a}_{< i} $, $ \bm{a}_0 $, and $ \bm{m} $, and we use $ \bm{f}^{(i)} $ to predict the conditional
distribution $ p(\bm{a}_{i} | \bm{a}_{< i}, \bm{m}, \bm{a}_0) $, which we describe next.
See our appendix for details.

\subsection{A Probabilistic Model of Actors}
\label{section:method/actor-model}

Having specified the generation process, we now turn our attention to modeling
each actor probabilistically.
As discussed earlier, each actor $ \bm{a}_i \in \mathcal{A} $
is parameterized by its class label $ c_i \in \mathbb{C} $,
location $ (x_i, y_i) \in \mathbb{R}^2 $,
oriented bounding box $ \bm{b}_i \in \mathbb{B} $
and velocity $ \bm{v}_i \in \mathbb{R}^2 $.
To capture the dependencies between these attributes, we factorize
$ p(\bm{a}_i | \bm{a}_{< i}, \bm{m}, \bm{a}_0) $ as follows:
\begin{align}
    p(\bm{a}_i) = p(c_i) p(x_i, y_i | c_i) p(\bm{b}_i | c_i, x_i, y_i) p(\bm{v}_i | c_i, x_i, y_i, \bm{b}_i)
\end{align}
where we dropped the condition on $ \bm{a}_{< i} $, $ \bm{m} $, and $ \bm{a}_0 $ to simplify notation.
Thus, the distribution over an actor's location is conditional on its class;
its bounding box is conditional on its class and location; and its velocity is
conditional on its class, location, and bounding box.
Note that if $ \bm{a}_i $ is the stopping token $ \bot $, we do not model its
location, bounding box, and velocity.
Instead, we have $ p(\bm{a}_i) = p(c_i) $, where $ c_i $ is the auxillary class $ c_\bot $.

\begin{figure*}
\includegraphics[width=\textwidth]{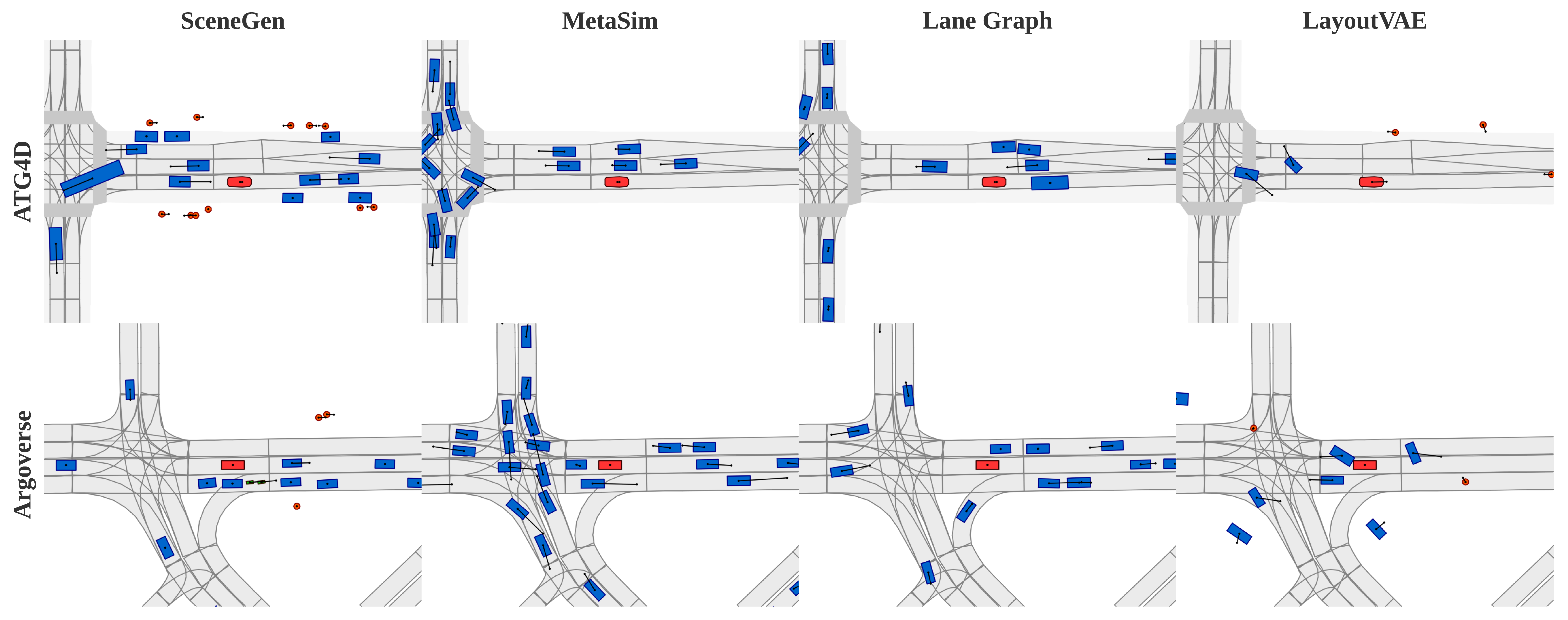}
\caption{
Qualitative comparison of traffic scenes generated by SceneGen and various baselines.
}
\label{figure:qualitative-comparison}
\end{figure*}

\paragraph{Class:}
To model a distribution over an actor's class, we use a categorical distribution
whose support is the set of classes $ \mathbb{C} \cup \{c_\bot\} $ and whose
parameters $ \bm{\pi}_\mathrm{c} $ are predicted by a neural network:
\begin{align}
    \bm{\pi}_\mathrm{c} &= \mathrm{MLP}_{\mathrm{c}}(\text{avg-pool}(\bm{f}^{(i)}); \bm{w}) \\
    c_i &\sim \mathrm{Categorical}(\bm{\pi}_\mathrm{c})
\end{align}
where $ \text{avg-pool} \colon \mathbb{R}^{C \times H \times W} \rightarrow \mathbb{R}^C $
is average pooling over the spatial dimensions and $ \mathrm{MLP}_{\mathrm{c}} $
is a three-layer multi-layer perceptron (MLP) with softmax activations.

\paragraph{Location:}
We apply uniform quantization to the actor's position and model the quantized
values using a categorical distribution.
The support of this distribution is the set of $ H \times W $ quantized bins
within our region of interest and its parameters $ \bm{\pi}_\mathrm{loc} $ are
predicted by a class-specific CNN.
This approach allows the model to express highly multi-modal distributions
without making assumptions about the distribution's shape~\cite{oord2016a}.
To recover continuous values, we assume a uniform distribution within each
quantization bin.

Let $ k $ denote an index into one of the $ H \times W $ quantized bins, and
suppose $ \lfloor\bm{p}_k\rfloor \in \mathbb{R}^2 $
(\emph{resp.}, $ \lceil\bm{p}_k\rceil \in \mathbb{R}^2 $)
is the minimum (\emph{resp.}, maximum) continuous coordinates in the $ k $-th bin.
We model $ p(x_i, y_i | c_i) $ as follows:
\begin{align}
    \bm{\pi}_{\mathrm{loc}} &= \mathrm{CNN}_{\mathrm{loc}}(\bm{f}^{(i)}; c_i, \bm{w}) \\
    k &\sim \mathrm{Categorical}(\bm{\pi}_\mathrm{loc}) \\
    (x_i, y_i) &\sim \mathrm{Uniform}(\lfloor \bm{p}_k \rfloor, \lceil \bm{p}_k \rceil)
\end{align}
where $ \mathrm{CNN}_\mathrm{loc}(\cdot; c_i, \bm{w}) $ is a CNN with softmax
activations for the class $ c_i $.
During inference, we mask and re-normalize $ \bm{\pi}_\mathrm{loc} $ such that
quantized bins with invalid positions according to our canonical ordering have
zero probability mass.
Note that we do not mask during training since this resulted in worse
performance.

After sampling the actor's location $ (x_i, y_i) \in \mathbb{R}^2 $, we extract
a feature vector $ \bm{f}^{(i)}_{x_i, y_i} \in \mathbb{R}^C $ by spatially
indexing into the $ k $-th bin of $ \bm{f}^{(i)} $.
This feature vector captures local information at $ (x_i, y_i) $ and is used
to subsequently predict the actor's bounding box and velocity.

\begin{table*}[!t]
\centering
\small
\begin{tabular}{l c c c c c c c c c c c c c c}
\toprule
              & \multicolumn{6}{c}{{\ourdataset}}                                                                 & \multicolumn{6}{c}{Argoverse} \\
                \cmidrule(lr){2-7}                                                                                         \cmidrule(lr){8-13}
Method        & NLL            & Features      & Class          & Size          & Speed         & Heading         & NLL            & Features      & Class         & Size          & Speed         & Heading \\
\midrule
Prob. Grammar & -              & 0.20          & 0.24           & 0.46          & 0.34          & 0.31            & -              & 0.38          & 0.26          & 0.41          & 0.57          & 0.38 \\
MetaSim       & -              & 0.12          & 0.24           & 0.45          & 0.35          & 0.15            & -              & 0.18          & 0.26          & 0.50          & 0.52          & 0.18 \\
Procedural    & -              & 0.38          & 0.24           & 0.17          & 0.34          & \textbf{0.07}   & -              & 0.58          & 0.26          & 0.23          & 0.59          & 0.17 \\
Lane Graph    & -              & 0.17          & 0.24           & 0.30          & 0.32          & 0.16            & -              & \textbf{0.11} & 0.26          & 0.31          & 0.32          & 0.28 \\
LayoutVAE     & 210.80         & 0.15          & \textbf{0.12}  & 0.18          & 0.33          & 0.29            & 200.78         & 0.25          & \textbf{0.11} & 0.21          & 0.41          & 0.29 \\
\midrule
SceneGen      & \textbf{59.86} & \textbf{0.11} & 0.20           & \textbf{0.06} & \textbf{0.27} & 0.08            & \textbf{67.11} & 0.14          & 0.21          & \textbf{0.17} & \textbf{0.17} & \textbf{0.21}\\
\bottomrule
\end{tabular}
\caption{
Negative log-likelihood (NLL) and maximum mean discrepency (MMD) results
on {\ourdataset} and Argoverse.
NLL is reported in \emph{nats}, averaged across all scenes in the test set.
MMD is computed between distributions of features extracted by a motion forecasting
model and various scene statistics (see main text for description).
For all metrics, lower is better.
}
\label{table:quantitative}
\end{table*}

\paragraph{Bounding box:}
An actor's bounding box $ \bm{b}_i \in \mathbb{B} $ consists of its width and
height $ (w_i, l_i) \in \mathbb{R}^2_{> 0} $ and its heading $ \theta_i \in [0, 2\pi) $.
We model the distributions over each of these independently.
For an actor's bounding box size, we use a mixture of
$ K $ bivariate log-normal distributions:
\begin{align}
    [\bm{\pi}_{\mathrm{box}}, \bm{\mu}_{\mathrm{box}}, \bm{\Sigma}_{\mathrm{box}}] &= \mathrm{MLP}_{\mathrm{box}}(\bm{f}^{(i)}_{x_i, y_i}; c_i, \bm{w}) \\
    k &\sim \mathrm{Categorical}(\bm{\pi}_{\mathrm{box}}) \\
    (w_i, l_i) &\sim \mathrm{LogNormal}(\bm{\mu}_{\mathrm{box}, k}, \bm{\Sigma}_{\mathrm{box}, k})
\end{align}
where $ \bm{\pi}_{\mathrm{box}} $ are mixture weights,
each $ \bm{\mu}_{\mathrm{box}, k} \in \mathbb{R}^2 $
and $ \bm{\Sigma}_{\mathrm{box}, k} \in \mathbb{S}_+^2 $ parameterize a
component log-normal distribution,
and $ \mathrm{MLP}_{\mathrm{box}}(\cdot; c_i, \bm{w}) $ is a three-layer MLP
for the class $ c_i $.
This parameterization allows our model to naturally capture the multi-modality
of actor sizes in real world data; \eg, the size of sedans versus trucks.

Similarly, we model an actor's heading angle with a mixture of
$ K $ Von-Mises distributions:
\begin{align}
    [\bm{\pi}_{\theta}, \mu_{\theta}, \kappa_{\theta}] &= \mathrm{MLP}_{\theta}(\bm{f}^{(i)}_{x_i, y_i}; c_i, \bm{w}) \\
    k &\sim \mathrm{Categorical}(\bm{\pi}_\theta) \\
    \theta_i &\sim \mathrm{VonMises}(\mu_{\theta, k}, \kappa_{\theta, k})
\end{align}
where $ \bm{\pi}_{\theta} $ are mixture weights,
each $ \mu_{\theta, k} \in [0, 2\pi) $ and
$ \kappa_{\theta, k} > 0 $ parameterize a component Von-Mises distribution,
and $ \mathrm{MLP}_{\theta}(\cdot; c_i, \bm{w}) $ is a three-layer MLP for
the class $ c_i $.
The Von-Mises distribution is a close approximation of a normal distribution
wrapped around the unit circle~\cite{prokudin2018} and has the probability
density function
\begin{align}
    p(\theta | \mu, \kappa) = \frac{e^{\kappa \cos(\theta - \mu)}}{2 \pi I_0(\kappa)}
\end{align}
where $ I_0 $ is the modified Bessel function of order 0.
We use a mixture of Von-Mises distributions to capture the
multi-modality of headings in real world data; \eg, a vehicle can
go straight or turn at an intersection.
To sample from a mixture of Von-Mises distributions, we sample a component
$ k $ from a categorical distribution and then sample $ \theta $ from the Von-Mises
distribution of the $ k $-th component~\cite{best1979}.

\paragraph{Velocity:}
We parameterize the actor's velocity $ \bm{v}_i \in \mathbb{R}^2 $ as
$ \bm{v}_i = (s_i \cos \omega_i, s_i \sin \omega_i) $, where
$ s_i \in \mathbb{R}_{\geq 0} $ is its speed and
$ \omega_i \in [0, 2\pi) $ is its direction.
Note that this parameterization is not unique since $ \omega_i $ can take any
value in $ [0, 2\pi) $ when $ \bm{v}_i = 0 $.
Therefore, we model the actor's velocity as a mixture model where one of the
$ K \geq 2 $ components corresponds to $ \bm{v}_i = 0 $.
More concretely, we have
\begin{align}
    \bm{\pi}_{\mathrm{v}} &= \mathrm{MLP}_{\mathrm{v}}(\bm{f}^{(i)}_{x_i, y_i}; c_i, \bm{w}) \\
    k &\sim \mathrm{Categorical}(\bm{\pi}_\mathrm{v})
\end{align}
where for $ k > 0 $, we have $ \bm{v}_i = (s_i \cos \omega_i, s_i \sin \omega_i) $, with
\begin{align}
    [\mu_{\mathrm{s}}, \sigma_{\mathrm{s}}] &= \mathrm{MLP}_{\mathrm{s}}(\bm{f}^{(i)}_{x_i, y_i}; c_i, \bm{w}) \\
    [\mu_{\omega}, \kappa_{\omega}] &= \mathrm{MLP}_{\omega}(\bm{f}^{(i)}_{x_i, y_i}; c_i, \bm{w}) \\
    s_i &\sim \mathrm{LogNormal}(\mu_{\mathrm{s}, k}, \sigma_{\mathrm{s}, k}) \\
    \omega_i &\sim \mathrm{VonMises}(\mu_{\mathrm{\omega}, k}, \kappa_{\mathrm{\omega}, k})
\end{align}
and for $ k = 0 $, we have $ \bm{v}_i = 0 $.
As before, we use three-layer MLPs to predict the parameters of each distribution.

For vehicles and bicyclists, we parameterize $ \omega_i \in [0, 2\pi) $
as an offset relative to the actor's heading $ \theta_i \in [0, 2\pi) $.
This is equivalent to parameterizing their velocities with
a bicycle model~\cite{taheri1990}, which we found improves sample quality.

\subsection{Learning and Inference}
\label{section:method/training}
\paragraph{Sampling:}
Pure sampling from deep autoregressive models can lead to degenerate examples
due to their ``unrealiable long tails''~\cite{holtzman2020}.
Therefore, we adopt a sampling strategy inspired by \emph{nucleus sampling}~\cite{holtzman2020}.
Specifically, at each generation step, we sample from each of
{\modelname}'s output distributions $ M $ times and keep the most likely sample.
We found this to help avoid degenerate traffic scenes while maintaining sample diversity.
Furthermore, we reject vehicles and bicyclists whose bounding boxes collide with
those of the actors sampled so far.

\paragraph{Training:}
We train our model to maximize the log-likelihood of real traffic scenes in
our training dataset:
\begin{align}
    \bm{w}^\star &= \mathrm{arg}\max_{\bm{w}} \sum_{i = 1}^{N} \log p(\bm{a}_{i, 1}, \ldots, \bm{a}_{i, n} | \bm{m}_i, \bm{a}_{i, 0}; \bm{w})
\end{align}
where $ \bm{w} $ are the neural network parameters and $ N $ is the number of
samples in our training set.
In practice, we use the Adam optimizer~\cite{kingma2015} to minimize the
average negative log-likelihood over mini-batches.
We use teacher forcing and backpropagation-through-time to train through the
generation process, up to a fixed window as memory allows.


\section{Experiments}
We evaluate SceneGen on two self-driving datasets:
Argoverse~\cite{argoverse2019} and {\ourdataset}~\cite{yang2018}.
Our results show that SceneGen can generate more realistic traffic
scenes than the competing methods (Sec.~\ref{section:experiments/results}).
We also demonstrate how {\modelname} with sensor simulation
can be used to train perception models that generalize to the real world
(Sec.~\ref{section:experiments/downstream}).

\subsection{Datasets}

\paragraph{{\ourdataset}:}
{\ourdataset}~\cite{yang2018} is a large-scale dataset collected by a fleet of SDVs
in cities across North America.
It consists of 5500 25-seconds logs which we split into a training set of
5000 and an evaluation set of 500.
Each log is subsampled at 10Hz to yield 250 traffic scenes, and each scene
is annotated with bounding boxes for vehicles, pedestrians, and
bicyclists.
Each log also provides HD maps that encode
lane boundaries, drivable areas, and crosswalks as polygons, and lane
centerlines as polylines.
Each lane segment is annotated with attributes such as its type (car \vs bike),
turn direction, boundary colors, and traffic light state.

In our experiments, we subdivide the training set into two splits of 4000
and 1000 logs respectively.
We use the first split to train the traffic scene generation models and the
second split to train the perception models in Sec.~\ref{section:experiments/downstream}.

\paragraph{Argoverse:}
Argoverse~\cite{argoverse2019} consists of two datasets
collected by a fleet of SDVs in Pittsburgh and Miami.
We use the Argoverse 3D Tracking dataset which contains
track annotations for 65 training logs and 24 validation logs.
Each log is subsampled at 10Hz to yield 13,122 training scenes
and 5015 validation scenes.
As in {\ourdataset}, Argoverse provides HD maps annotated with drivable areas
and lane segment centerlines and their attributes; \eg, turn direction.
However, Argoverse does not provide crosswalk polygons,
lane types, lane boundary colors, and traffic lights.

\begin{table}[!t]
\centering
\small
\begin{tabular}{c c c c c}
\toprule
\# Mixtures & Scene          & Vehicle       & Pedestrian    & Bicyclist \\
\midrule
1           & 125.97         & 7.26          & 10.36         & 9.16 \\
3           & 68.22          & 2.64          & 8.52          & 7.34 \\
5           & 64.05          & 2.35          & \textbf{8.27} & 7.22 \\
10          & \textbf{59.86} & \textbf{1.94} & 8.32          & \textbf{6.90} \\
\bottomrule
\end{tabular}
\caption{
Ablation of the number of mixture components in {\ourdataset}.
Scene NLL is averaged across scenes and NLL per class is the average
NLL per actor of that class.
}
\label{table:ablation-modes}
\end{table}

\begin{table}[!t]
\centering
\small
\begin{tabular}{c c c c c c c c}
\toprule
L          & DA         & C          & TL          & Scene          & Veh.          & Ped.          & Bic. \\
\midrule
           &            &            &             & 93.73          & 4.90          & 8.85          & 7.17 \\
\checkmark &            &            &             & 63.33          & 2.12          & 8.69          & 7.10 \\
\checkmark & \checkmark &            &             & \textbf{57.66} & \textbf{1.73} & 8.40          & 6.84 \\
\checkmark & \checkmark & \checkmark &             & 57.96          & 1.77          & \textbf{8.32} & \textbf{6.61} \\
\checkmark & \checkmark & \checkmark & \checkmark  & 59.86          & 1.94          & \textbf{8.32} & 6.90 \\
\bottomrule
\end{tabular}
\caption{
Ablation of map in {\ourdataset} (in NLL).
\textbf{L} is lanes;
\textbf{DA} drivable areas;
\textbf{C} crosswalks;
and \textbf{TL} traffic lights.
}
\label{table:ablation-map}
\end{table}

\subsection{Experiment Setup}

\paragraph{Baselines:}
Our first set of baselines is inspired by recent work on probabilistic scene
grammars and graphs~\cite{prakash2019,kar2019,devaranjan2020}.
In particular, we design a probabilistic grammar of traffic scenes (\textbf{Prob. Grammar})
such that actors are randomly placed onto lane segments using a
hand-crafted prior~\cite{prakash2019}.
Sampling from this grammar yields a \emph{scene graph}, and
our next baseline (\textbf{MetaSim}) uses a graph neural network to transform
the attributes of each actor in the scene graph.
Our implementation follows Kar~\etal~\cite{kar2019}, except we use a training
algorithm that is supervised with heuristically generated ground truth scene
graphs.\footnote{We were unable to train MetaSim using their unsupervised losses.}

Our next set of baselines is inspired by methods that reason directly about
the road topology of the scene~\cite{wheeler2015,wheeler2016,jesenski2019,manivasagam2020}.
Given a \emph{lane graph} of the scene, \textbf{Procedural} uses
a set of rules to place actors such that they follow lane centerlines, maintain
a minimum clearance to leading actors, \etc.
Each actor's bounding box is sampled from a Gaussian KDE fitted to the
training dataset~\cite{bishop2006} and velocities are set to satisfy speed
limits and a time gap between successive actors on the lane graph.
Similar to MetaSim, we also consider a learning-based version of Procedural
that uses a lane graph neural network~\cite{liang2020} to transform each
actor's position, bounding box, and velocity (\textbf{Lane Graph}).

Since the HD maps in {\ourdataset} and Argoverse do not provide reference
paths for pedestrians, the aforementioned baselines cannot generate pedestrians.\footnote{
In Argoverse, these baselines generate vehicles only since bike lanes are not
given.
This highlights the challenge of designing good heuristics.}
Therefore, we also compare against \textbf{LayoutVAE}~\cite{jyothi2019}---a
variational autoencoder for image layouts that we adapt for traffic
scene generation.
We modify LayoutVAE to condition on HD maps and output oriented
bounding boxes and velocities for actors of every class.
Please see our appendix for details.

\paragraph{Metrics:}
Our first metric measures the negative log-likelihood (\textbf{NLL}) of real
traffic scenes from the evaluation distribution, measured in \emph{nats}.
NLL is a standard metric to compare generative models with tractable
likelihoods.
However, as many of our baselines do not have likelihoods, we
compute a sample-based metric as well: maximum mean discrepancy (\textbf{MMD})~\cite{gretton2012}.
For two distributions $ p $ and $ q $, MMD measures a distance between $ p $ and $ q $ as
\begin{equation}
\begin{split}
    \mathrm{MMD}^2(p, q) = \mathbb{E}_{x, x'\sim p}[k(x, x')] \\
    + \mathbb{E}_{y, y' \sim q}[k(y, y')]
    - 2\mathbb{E}_{x\sim p, y \sim q}[k(x, y)]
\end{split}
\end{equation}
for some kernel $ k $.
Following \cite{you2018,liao2019}, we compute MMD using Gaussian kernels with
the total variation distance to compare distributions of scene statistics
between generated and real traffic scenes.
Our scene statistics measure the distribution of classes, bounding box sizes,
speeds, and heading angles (relative to the SDV) in each scene.
To peer into the global properties of the traffic scene,
we also compute MMD in the feature space of a pre-trained motion forecasting
model that takes a rasterized image of the scene as input~\cite{wong2020}.
This is akin to the popular IS~\cite{salimans2016}, FID~\cite{heusel2017}, and
KID~\cite{binkowski2018} metrics for evaluating generative models, except we
use a feature extractor trained on traffic scenes.
Please see our appendix for details.

\paragraph{Additional details:}
Each traffic scene is a $ 80m \times 80m $ region of interest centered on the ego SDV.
By default, {\modelname} uses $ K = 10 $ mixture components and
conditions on all available map elements for each dataset.
We train {\modelname} using the Adam optimizer~\cite{kingma2015} with a learning
rate of $ 1\mathrm{e-}4 $ and a batch size of 16, until convergence.
When sampling each actor's position, heading, and velocity, we sample
$ M = 10 $ times and keep the most likely sample.

\subsection{Results}
\label{section:experiments/results}

\paragraph{Quantitative results:}
Tab.~\ref{table:quantitative} summarizes the NLL and MMD results for {\ourdataset} and Argoverse.
Overall, {\modelname} achieves the best results across both datasets, demonstrating
that it can better model real traffic scenes and synthesize realistic examples as well.
Interestingly, all learning-based methods outperform the hand-tuned baselines
with respect to MMD on deep features---a testament to the difficulty of designing good heuristics.

\paragraph{Qualitative results:}
Fig.~\ref{figure:qualitative-results} visualizes samples generated by
{\modelname} on {\ourdataset} and Argoverse.
Fig.~\ref{figure:qualitative-comparison} compares
traffic scenes generated by {\modelname} and various baselines.
Although MetaSim and Lane Graph generate reasonable scenes, they are
limited by their underlying heuristics; \eg, actors follow lane centerlines.
LayoutVAE generates a greater variety of actors; however, the model
does not position actors on the map accurately, rendering the overall
scene unrealistic.
In contrast, {\modelname}'s samples reflects the complexity of real traffic
scenes much better.
That said, {\modelname} occassionally generates near-collision scenes that
are plausible but unlikely; \eg, Fig.~\ref{figure:qualitative-results} top-right.

\begin{figure}
\includegraphics[width=\linewidth]{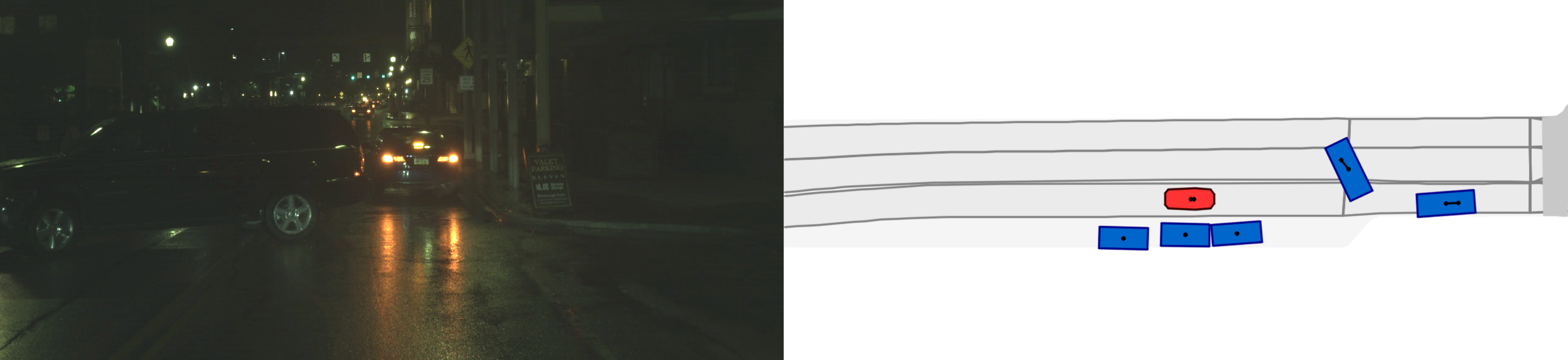}
\caption{{\ourdataset} scene with a traffic violation.}
\label{figure:nll-mining}
\end{figure}

\paragraph{Ablation studies:}
In Tab.~\ref{table:ablation-modes}, we sweep over the number of mixture
components used to parameterize distributions of bounding boxes and velocities.
We see that increasing the number of components consistently lowers NLL,
reflecting the need to model the multi-modality of real traffic scenes.
We also ablate the input map to {\modelname}:
starting from an unconditional model, we progressively add
lanes, drivable areas, crosswalks, and traffic light states.
From Tab.~\ref{table:ablation-map}, we see that using more map
elements generally improves NLL.
Surprisingly, incorporating traffic lights slightly degrades performance,
which we conjecture is due to infrequent traffic light observations
in {\ourdataset}.

\paragraph{Discovering interesting scenes:}
We use {\modelname} to find unlikely scenes in {\ourdataset} by searching for
scenes with the highest NLL, normalized by the number of actors.
Fig.~\ref{figure:nll-mining} shows an example of a traffic violation found via
this procedure; the violating actor has an NLL of 21.28.

\begin{table}[!t]
\centering
\small
\begin{tabular}{l c c c c c c }
\toprule
               & \multicolumn{2}{c}{Vehicle} & \multicolumn{2}{c}{Pedestrian} & \multicolumn{2}{c}{Bicyclist} \\
Method         & 0.5           & 0.7           & 0.3           & 0.5            & 0.3           & 0.5           \\
\midrule
Prob. Gram.    & 81.1          & 66.6          & -             & -              & 11.2          & 10.6          \\
MetaSim        & 76.3          & 63.3          & -             & -              & 8.2           & 7.5           \\
Procedural     & 80.2          & 63.0          & -             & -              & 6.5           & 3.8           \\
Lane Graph     & 82.9          & 71.7          & -             & -              & 7.6           & 6.9           \\
LayoutVAE      & 85.9          & 76.3          & 49.3          & 41.8           & 18.4          & 16.4          \\
SceneGen       & \textbf{90.4} & \textbf{82.4} & \textbf{58.1} & \textbf{48.7}  & \textbf{19.6} & \textbf{17.9} \\
\midrule
Real Scenes    & 93.7          & 86.7          & 69.3          & 61.6           & 29.2          & 25.9 \\
\bottomrule
\end{tabular}
\caption{
Detection AP on real {\ourdataset} scenes.
}
\label{table:data-augmentation}
\end{table}

\begin{figure}
\includegraphics[width=0.49\linewidth]{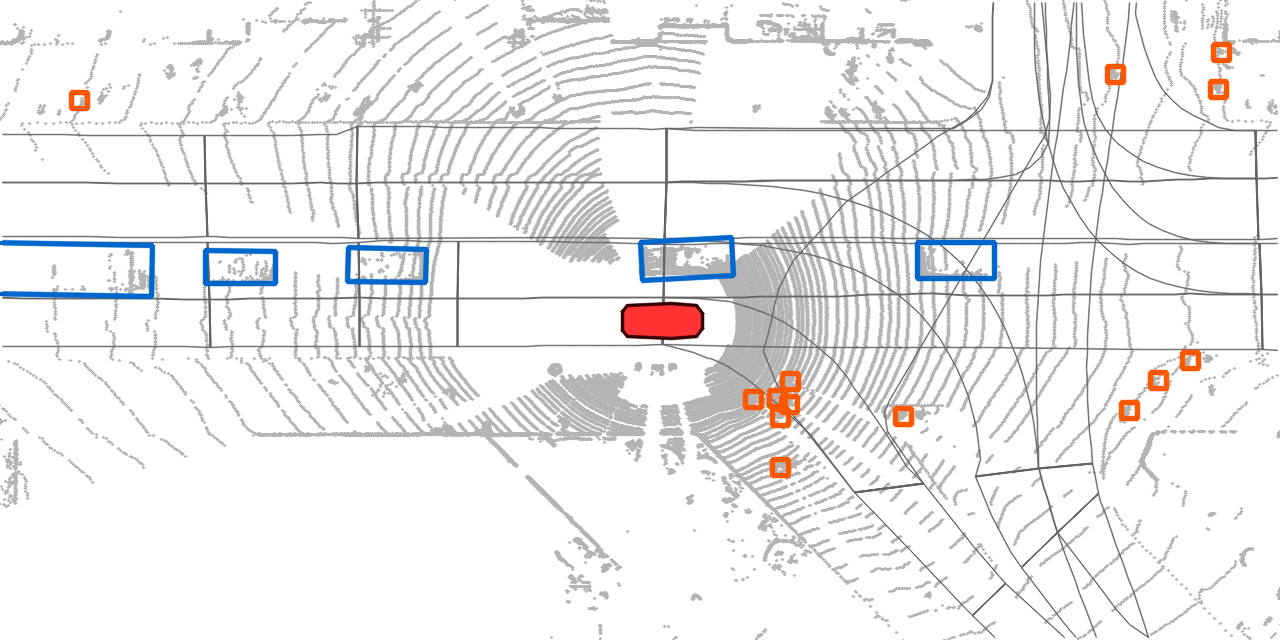}
\includegraphics[width=0.49\linewidth]{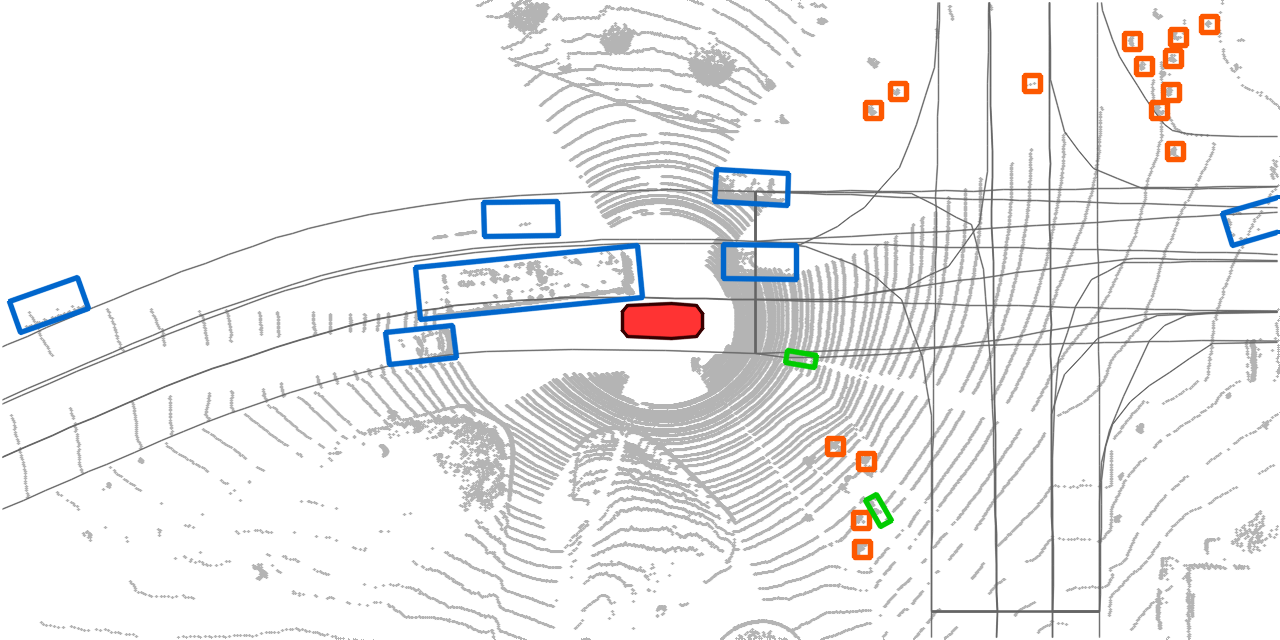}
\caption{Outputs of detector trained with SceneGen scenes.}
\label{figure:downstream-qualitative-results}
\end{figure}

\subsection{Sim2Real Evaluation}
\label{section:experiments/downstream}

Our next experiment demonstrates that {\modelname} coupled with sensor
simulation can generate realistic labeled data for training perception models.
For each method under evaluation, we generate 250,000 traffic scenes conditioned on the
SDV and HD map in each frame of the 1000 held-out logs in {\ourdataset}.
Next, we use LiDARsim~\cite{manivasagam2020} to simulate the LiDAR point cloud
corresponding to each scene.
Finally, we train a 3D object detector~\cite{yang2018} using the simulated LiDAR
and evaluate its performance on real scenes and LiDAR in {\ourdataset}.

From Tab.~\ref{table:data-augmentation}, we see that {\modelname}'s traffic
scenes exhibit the lowest sim2real gap.
Here, Real Scenes is simulated LiDAR from ground truth placements.
This reaffirms our claim that the underlying rules and priors used in
MetaSim and Lane Graph induce a content gap.
By eschewing these heuristics altogether, {\modelname} learns to
generate significantly more realistic traffic scenes.
Intriguingly, LayoutVAE performs competitively despite
struggling to position actors on the map.
We conjecture that this is because LayoutVAE captures the diversity
of actor classes, sizes, headings, \etc well.
However, by accurately modeling actor positions as well, {\modelname}
further reduces the sim2real gap, as compared to ground truth traffic scenes.

\section{Conclusion}
We have presented {\modelname}---a neural autoregressive model of traffic scenes
from which we can sample new examples as well as evaluate the likelihood of
existing ones.
Unlike prior methods, {\modelname} eschews the need for rules or heuristics,
making it a more flexible and scalable approach for modeling the complexity
and diversity of real world traffic scenes.
As a result, {\modelname} is able to generate realistic traffic scenes,
thus taking an important step towards safe and scalable self-driving.

{\small
\bibliographystyle{ieee_fullname}
\bibliography{egbib}
}

\end{document}


\title{Supplementary Materials \\
SceneGen: Learning to Generate Realistic Traffic Scenes}

\author{
Shuhan Tan$^{1,2}$\thanks{Indicates equal contribution. Work done at Uber ATG.}
\quad Kelvin Wong$^{1,3*}$ \quad Shenlong Wang$^{1,3}$ \\
Sivabalan Manivasagam$^{1,3}$ \quad Mengye Ren$^{1,3}$ \quad Raquel Urtasun$^{1,3}$ \\
$^{1}$Uber Advanced Technologies Group \quad $^{2}$Sun Yat-Sen University \quad $^{3}$University of Toronto \\
{
    \tt\small{tanshh@mail2.sysu.edu.cn} \quad
    \tt\small{\{kelvinwong,slwang,manivasagam,mren,urtasun\}@cs.toronto.edu}
}
}

\maketitle

\begin{abstract}
In our supplementary materials, we detail {\modelname}'s model architecture
and training procedure (Sec.~\ref{section:method}).
Additionally, we provide additional experiment details in Sec.~\ref{section:experiments}
and additional experiment results in Sec.~\ref{section:quantiative}.
In Sec.~\ref{section:qualitative}, we exhibit an extensive array of
qualitative results that demonstrate the realism and diversity of the traffic
scenes generated by {\modelname}.
\end{abstract}

\section{Additional Model Details}
\label{section:method}

\subsection{Input Representation}
\label{section:method/input-representation}
At each step of the generation process, {\modelname} takes a bird's eye
view multi-channel image encoding the HD map $ \bm{m} $, the SDV $ \bm{a}_0 $,
and the actors generated so far $ \{\bm{a}_1, \ldots, \bm{a}_{i - 1}\} $.
The image emcompasses an $ 80m \times 80m $ region of interest centered on
the SDV $ \bm{a}_0 $ and has a resolution of $ 0.25m $ per pixel,
yielding a $ 320 \times 320 $ image.
The HD map is rasterized into a multi-channel image describing all available map
elements in each dataset.
For {\ourdataset}, our multi-channel image consists of:
lane polygons (straight vehicle lanes, dedicated right vehicle lanes,
dedicated left vehicle lanes, dedicated bus lanes, and dedicated bike lanes);
lane centerlines and dividers (allowed to cross, forbidden to cross, and maybe
allowed to cross);
lane segments (straight vehicle lanes, dedicated right vehicle lanes, and dedicated left vehicle lanes);
drivable area and road polygons;
and crosswalk polygons.
In addition, we also encode each lane segment's traffic light state
(green, yellow, red, flashing yellow, flashing red, and unknown),
speed limit, and orientation as filled lane polygons.
Note that orientation angles are encoded in their Biternion representations
$ \theta = (\cos \theta, \sin \theta) $~\cite{prokudin2018}.
In aggregate, this yields a 24-channel image.

Argoverse provides a more limited set of map elements.
Here, our multi-channel image consists of:
lane polygons;
lane centerlines (all lanes, left turn lanes, right turn lanes,
intersection lanes, and traffic-controlled lanes);
lane orientations (in Biternion representation); and
drivable area polygons.
In aggregate, this yields a 9-channel image.

\begin{figure*}[t]
\includegraphics[width=\textwidth]{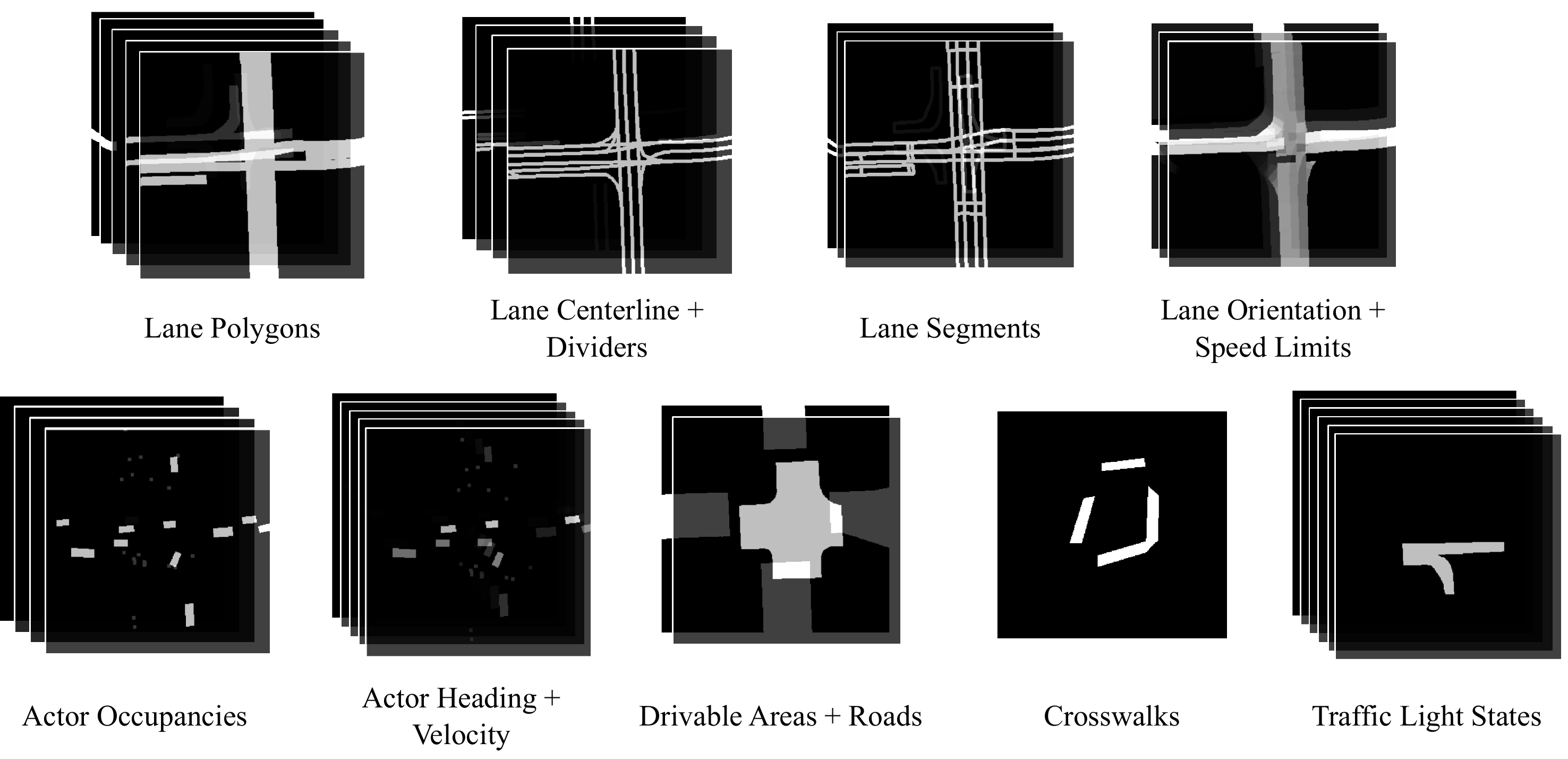}
\caption{The input multi-channel image to {\modelname} for {\ourdataset}.}
\label{figure:rasterization}
\end{figure*}

To encode the actors $ \bm{a}_0, \bm{a}_1, \ldots $, we rasterize their
bounding boxes onto a collection of binary occupancy images~\cite{bansal2019},
one for each class; \ie, SDV, vehicles, pedestrians, and bicyclists.
Furthermore, we encode their headings and velocities by rasterizing their
bounding boxes onto a five-channel image, filled with their respective speed,
direction, and heading.
As before, direction and heading angles are encoded in their Biternion representations.
See Fig.~\ref{figure:rasterization} for an example.

\subsection{Model Architecture}
The basis of our model is the ConvLSTM architecture~\cite{shi2015}.
Let $ \bm{x}^{(i)} \in \mathbb{R}^{C \times H \times W} $ denote the input
multi-channel image at the $ i $-th step of the generation process.
Given the previous hidden and cell states $ \bm{h}^{(i - 1)} $ and $ \bm{c}^{(i - 1)} $,
the new hidden states $ \bm{h}^{(i)} $, cell states $ \bm{c}^{(i)} $, and
backbone features $ \bm{f}^{(i)} $ are given by:
\begin{align}
    \bm{h}^{(i)}, \bm{c}^{(i)} &= \mathrm{ConvLSTM}(\bm{x}^{(i)}, \bm{h}^{(i - 1)}, \bm{c}^{(i - 1)}; \bm{w}) \\
    \bm{f}^{(i)} &= \mathrm{CNN}_\mathrm{b}(\bm{h}^{(i)}; \bm{w})
\end{align}

Here, ConvLSTM is a two-layer ConvLSTM with
$ 5 \times 5 $ convolution kernels and 32 hidden channels,
and $ \mathrm{CNN}_\mathrm{b} $ is a five-layer convolutional neural network (CNN)
with 32 feature channels per layer.
Each convolution layer consists of a $ 3 \times 3 $ convolution kernel,
Group Normalization~\cite{wu2018}, and ReLU activations.
The backbone features $ \bm{f}^{(i)} $ summarize the generated scene so far
and are given as input to the subsequent actor modules, which we detail next.

\paragraph{Class:}
We predict the class categorical distribution parameters
$ \bm{\pi}_c \in \Delta^{|\mathbb{C}|} $ as follows\footnote{
We use $ \Delta^{n} = \{(x_0, x_1, \ldots, x_n) \in \mathbb{R}^{n + 1} | \sum_i x_i = 1 \text{ and } x_i \geq 0 \text{ for all } i\} $ to denote the $ n $-simplex.}:
\begin{align}
    \bm{\pi}_c = \mathrm{MLP}_\mathrm{c}(\text{avg-pool}(\bm{f}^{(i)}); \bm{w})
\end{align}
where $ \text{avg-pool} \colon \mathbb{R}^{C \times H \times W} \rightarrow \mathbb{R}^C $
is average pooling over the spatial dimensions and $ \mathrm{MLP}_\mathrm{c} $
is a three-layer multi-layer perceptron (MLP) with 32 feature channels per
hidden layer, ReLU activations, and softmax outputs.

\paragraph{Location:}
We apply uniform quantization to each actor's position and model the quantized
values with a categorical distribution.
Our quantization resolution is $ 0.25m $, which we found sufficient to generate
realistic traffic scenes while balancing memory efficiency.
To predict the parameters $ \bm{\pi}_\mathrm{loc} \in \Delta^{H \times W - 1} $,
we use a three-layer CNN with 32 feature channels per hidden layer.
Each hidden convolution layer consists of a $ 3 \times 3 $ convolution kernel,
Group Normalization~\cite{wu2018}, and ReLU activations.
The output convolution layer uses a $ 1 \times 1 $ kernel with softmax activations.
Note that we use separate CNN weights for each class in $ \mathbb{C} $; \ie,
vehicles, pedestrians, and bicyclists.

\paragraph{Bounding box:}
An actor's bounding box $ \bm{b}_i \in \mathbb{B} $ consists of its width
and height $ (w_i, l_i) \in \mathbb{R}^2_{> 0} $ and its heading $ \theta_i \in [0, 2\pi) $.
We model the distribution over bounding box sizes with a mixture of $ K $
bivariate log-normal distributions whose parameters are predicted by a
three-layer MLP (with the same architecture as described earlier):
\begin{align}
    [\bm{\pi}_{\mathrm{box}}, \bm{\mu}_{\mathrm{box}}, \bm{\Sigma}_{\mathrm{box}}] &= \mathrm{MLP}_{\mathrm{box}}(\bm{f}^{(i)}_{x_i, y_i}; c_i, \bm{w})
\end{align}
where $ \bm{\pi}_{\mathrm{box}} \in \Delta^{K - 1} $ are mixture weights and each $ \bm{\mu}_{\mathrm{box}, k} \in \mathbb{R}^2 $
and $ \bm{\Sigma}_{\mathrm{box}, k} \in \mathbb{S}_+^2 $ parameterize a component log-normal distribution.
To enforce the constraint that each $ \bm{\Sigma} \in \mathbb{S}_+^2 $,
$ \mathrm{MLP}_\mathrm{box} $ predicts a variance term $ \bm{\sigma}^2 \in \mathbb{R}^2_{> 0} $ (in log-scale)
and a correlation term $ \rho \in [-1, 1] $ (using tanh) such that:
\begin{align}
\bm{\Sigma} = \begin{bmatrix}
    \sigma^2_{1}  & \rho \sigma_{1} \sigma_{2} \\
    \rho \sigma_{1} \sigma_{2} & \sigma^2_{2}
\end{bmatrix} \in \mathbb{S}_+^2
\end{align}

Similarly, we model the distribution over heading angles with a mixture
of $ K $ Von-Mises distributions whose parameters are predicted by another
three-layer MLP:
\begin{align}
    [\bm{\pi}_{\theta}, \mu_{\theta}, \kappa_{\theta}] &= \mathrm{MLP}_{\theta}(\bm{f}^{(i)}_{x_i, y_i}; c_i, \bm{w})
\end{align}
where $ \bm{\pi}_{\theta} \in \Delta^{K - 1} $ are mixture weights and each $ \mu_{\theta, k} \in [0, 2\pi) $ and
$ \kappa_{\theta, k} > 0 $ parameterize a component Von-Mises distribution.
Following Prokudin \etal~\cite{prokudin2018}, we parameterize each $ \mu $
with its Biternion representation $ \mu = (\cos \mu, \sin \mu) $ and each
$ \kappa $ is predicted in log-scale.
Note that we use separate MLP weights for each class in $ \mathbb{C} $ whose
actors are represented by bounding boxes; \ie, vehicles and bicyclists.
Pedestrians are represented by their center of gravity only (\ie, location).

\paragraph{Velocity:}
Each of $ \mathrm{MLP}_\mathrm{v} $, $ \mathrm{MLP}_s $, and $ \mathrm{MLP}_\omega $
is a three-layer MLP with the same architecture as described above.
We parameterize the mixture of $ K $ Von-Mises distributions for directions
$ \omega $ just as we parameterize the distribution of headings.
As before, we use separate MLP weights for each class in $ \mathbb{C} $.

\subsection{Training Details}
We train our model to maximize the log-likelihood of real traffic scenes in
our training dataset:
\begin{align}
    \bm{w}^\star &= \mathrm{arg}\max_{\bm{w}} \sum_{i = 1}^{N} \log p(\bm{a}_{i, 1}, \ldots, \bm{a}_{i, n} | \bm{m}_i, \bm{a}_{i, 0}; \bm{w})
\end{align}
where $ \bm{w} $ are the neural network parameters and $ N $ is the number of
samples in our training set.
We use teacher forcing and backpropagation-through-time to train through the
generation process, up to a fixed window as memory allows.
On a Nvidia Quadro RTX 5000 with 16GB of GPU memory, we train through 25
generation steps with batch size of 1 per GPU.
We use PyTorch~\cite{pytorch2019} and Horovod~\cite{horovod2018} to distribute
the training process over 16 GPUs with a total batch size of 16.
During training, we also randomly rotate each traffic scene with $ \theta \in [0, 2\pi) $.

Note that each summand $ \log p(\bm{a}_{1}, \ldots, \bm{a}_{n} | \bm{m}; \bm{w}) $ can be
decomposed into a sum of the log-likelihoods for each actors; namely, we have
\begin{align}
    \begin{split}
    \log p(\bm{a}_i | \bm{\xi}_i) = \underbrace{\log p(c_i | \bm{\xi}_i)}_{\text{class}}
      + \underbrace{\log p(x_i, y_i | c_i, \bm{\xi}_i)}_{\text{location}}
      + \underbrace{\log p (\bm{b}_i | c_i, x_i, y_i, \bm{\xi}_i)}_{\text{bounding box}}
      + \underbrace{\log p (\bm{v}_i | c_i, x_i, y_i, \bm{b}_i, \bm{\xi}_i)}_{\text{velocity}}
    \end{split}
\end{align}
where $ \bm{\xi}_i $ encapsulates the conditions on $ \bm{a}_{< i} $, $ \bm{m} $,
and $ \bm{a}_0 $, to simplify notation.
Therefore, the first summand $ \log p(c_i | \bm{\xi}_i) $ is the (negative) cross-entropy loss
between the predicted parameters $ \bm{\pi}_\mathrm{c} $ and the ground truth
class $ c_i \in \mathbb{C} \cup \{\bot\} $.
We describe the remaining summands in detail next.

\paragraph{Location:} The second summand $ \log p(x_i, y_i | c_i, \bm{\xi}_i) $ measures
the log-likelihood the actor's location $ (x_i, y_i) \in \mathbb{R}^2 $.
As discussed earlier, we uniformly quantize each actor's location and
parameterize it with a categorical distribution.
Therefore, $ \log p(x_i, y_i | c_i, \bm{\xi}_i) $ is the (negative) cross-entropy loss
between the predicted parameters $ \bm{\pi}_\mathrm{loc} $ and the actor's
ground truth quantized location.
To address the significant imbalance of positive versus negative locations here,
we use online negative hard mining.
Specifically, we normalize $ \bm{\pi}_\mathrm{loc} $ over the hardest 10,000
locations only (including the positive location), and compute
$ \log p(x_i, y_i | c_i, \bm{\xi}_i) $ based this restricted categorical
distribution instead.

\paragraph{Bounding box:}
The third summand $ \log p(\bm{b}_i | c_i, x_i, y_i, \bm{\xi}_i) $ is a sum of the
log-likelihoods of the actor's bounding box size $ (w_i, l_i) \in \mathbb{R}^2 $ and heading $ \theta_i \in [0, 2\pi) $:
\begin{align}
    \begin{split}
    \log p(\bm{b}_i | c_i, x_i, y_i, \bm{\xi}_i) = \log p(w_i, l_i | c_i, x_i, y_i, \bm{\xi}_i)
        + \log p(\theta_i | c_i, x_i, y_i, \bm{\xi}_i)
    \end{split}
\end{align}

Since we model bounding box size with a mixture of $ K $ bivariate log-normal
distributions, we have
\begin{align}
    \begin{split}
    \log p(w_i, l_i | c_i, x_i, y_i, \bm{\xi}_i) =
    \log \sum_{k = 1}^{K}
    \pi_k
    \frac{1}{2 \pi \sigma_{k, 1} \sigma_{k, 2} \sqrt{1 - \rho_k^2}}
    e^{-\frac{1}{2 (1 - \rho_k^2)}
    \left[
        \left(\frac{\log w_i - \mu_{k, 1}}{\sigma_{k, 1}}\right)^2
        + \left(\frac{\log l_i - \mu_{k, 2}}{\sigma_{k, 2}}\right)^2
        + 2 \rho_k \left(\frac{\log w_i - \mu_{k, 1}}{\sigma_{k, 1}}\right) \left(\frac{\log l_i - \mu_{k, 2}}{\sigma_{k, 2}}\right)
    \right]}
    \end{split}
\end{align}
where $ \bm{\pi} \in \Delta^{K - 1} $ are mixture weights and each
$ \bm{\mu}_k \in \mathbb{R}^2 $,
$ \bm{\sigma}_k \in \mathbb{R}^2_{> 0} $,
and $ \rho_k \in [-1, 1] $ parameterize a component bivariate log-normal distribution.

Similarly, since we model heading angles with a mixture of $ K $ Von-Mises distributions, we have
\begin{align}
    \begin{split}
    \log p(\theta_i | c_i, x_i, y_i, \bm{\xi}_i) =
    \log \sum_{k = 1}^{K}
    \pi_k
    \frac{e^{\kappa_k \cos (\theta_i - \mu_k)}}{2 \pi I_0(\kappa_k)}
    \end{split}
\end{align}
where $ \bm{\pi} \in \Delta^{K - 1} $ are mixture weights and each
$ \bm{\mu}_k \in [0, 2\pi) $ and $ \kappa_k > 0 $ parameterize a component
Von-Mises distribution.

\paragraph{Velocity:}
The fourth summand $ \log p(\bm{v}_i | c_i, x_i, y_i, \bm{b}_i, \bm{\xi}_i) $
is the log-likelihood of the actor's velocity $ \bm{v}_i \in \mathbb{R}^2 $,
which we parameterize as $ \bm{v}_i = (s_i \cos \omega_i, s_i \sin \omega_i) $
where $ s_i \in \mathbb{R}_{\geq 0} $ is its speed and $ \omega_i \in [0, 2\pi) $
is its direction.
Recall that we model the distribution over an actor's velocity as a mixture model
where one of the $ K \geq 2 $ components corresponds to $ \bm{v}_i = 0 $.
Therefore, for $ \bm{v}_i = 0 $, we have
\begin{align}
    \log p(\bm{v}_i | c_i, x_i, y_i, \bm{b}_i, \bm{\xi}_i) = \log \pi_1
\end{align}
and for $ \bm{v}_i > 0 $, we have
\begin{align}
    \log p(\bm{v}_i | c_i, x_i, y_i, \bm{b}_i, \bm{\xi}_i) =
    \log \sum_{k = 2}^{K}
    \pi_k
    \underbrace{\frac{1}{\sigma_{s, k} \sqrt{2 \pi}} e^{-\frac{1}{2} \left(\frac{\log s_i - \mu_{s, k}}{\sigma_{s, k}}\right)^2}}_{\text{speed}}
    \times
    \underbrace{\frac{e^{\kappa_{\omega, k} \cos(\omega_i - \mu_{\omega, k})}}{2 \pi I_0(\kappa_{\omega, k})}}_{\text{direction}}
\end{align}
where $ \bm{\pi} \in \Delta^{K - 1} $ are mixture weights,
each $ \mu_{s, k} \in \mathbb{R} $ and $ \sigma_{s, k} > 0 $ parameterize a
component log-normal distribution for speed $ s_i $,
and each $ \mu_{\omega, k} \in [0, 2\pi) $ and $ \kappa_{\omega, k} > 0 $
parameterize a component Von-Mises distribution for direction $ \omega_i $.

\section{Additional Experiment Details}
\label{section:experiments}

\subsection{Baselines}

\paragraph{Prob. Grammar:}
Our Prob. Grammar baseline is inspired by recent work on probabilistic scene
grammars~\cite{prakash2019,kar2019,devaranjan2020}.
Here, traffic scenes are composed by placing actors onto lane segments in the
HD map, and initializing their classes, sizes, headings, velocities according
to a hand-crafted prior.
In our experiments, we use the following scene grammar:
\begin{align}
    \langle\text{Scene}\rangle & \rightarrow \langle\text{Lanes}\rangle \\
    \langle\text{Lanes}\rangle & \rightarrow \langle\text{Lane}\rangle \langle\text{Lanes}\rangle | \epsilon \\
    \langle\text{Lane}\rangle  & \rightarrow \langle\text{Actors}\rangle \\
    \langle\text{Actors}\rangle & \rightarrow \langle\text{Actor}\rangle \langle\text{Actors}\rangle | \epsilon
\end{align}
where Actor and $ \epsilon $ are terminal symbols.
Sampling from this scene grammar yields a \emph{scene graph}, which
defines the \emph{scene structure}---where lane segments are and which actors
are positioned on top of them---and \emph{scene parameters}---the attributes
of each lane segment and actor.
In our setting, we are given the lane nodes (and the SDV actor's node) of the scene
graph as a condition, and our goal is to insert/modify the actor nodes.

Drawing inspiration from MetaSim's probabilistic scene grammar~\cite{kar2019},
we first uniformly sample the maximum number of actors per lane segment and then
place them along the lane centerline, with a random clearance between successive
actors drawn from the exponential distribution.
The class of each actor is determined by the lane segment under consideration
(\ie, car lane \vs bike lane);
its lateral offset from the lane centerline is given by uniform noise;
its bounding box size is sampled from a uniform distribution;
its heading and the direction of its velocity is given by the
direction of the lane segment plus some uniform noise;
and its speed is the minimum of a sample from a uniform distribution
and the lane segment's speed limit.
The parameters of every distribution are tuned by hand.

\paragraph{MetaSim:}
Our next baseline (MetaSim) uses a graph neural network (GNN) to transform
the attributes of each actor node in the given scene graph.
We use the implementation of Kar~\etal~\cite{kar2019} for this purpose.
Specifically, given a scene graph drawn from Prob. Grammar, MetaSim
deterministically modifies each actor's distance along its lane centerline,
lateral offset, bounding box size, heading, and velocity.
The inputs to MetaSim is a scene graph where each node's features are its
attributes (normalized between 0 and 1 based on their respective minimum/maximum
values under the prior), and the outputs of MetaSim are each node's new
attributes (again normalized between 0 and 1).
We use the GNN architecture of Kar~\etal~\cite{kar2019}: a three-layer GNN
encoder with $ 32 \rightarrow 64 \rightarrow 128 $ features channels and a
three-layer GNN decoder with $ 128 \rightarrow 64 \rightarrow 32 $ feature channels.
Additionally, we use linear layers to encode and decode the per-node attributes.

Note that we train MetaSim using a supervised algorithm with heuristically
generated ground truth scene graphs.
In particular, given a real traffic scene, we first associate each actor to a
lane segment; if this is not possible, the actor is not included in the scene graph.
Next, we modify the attributes of each actor according to Prob. Grammar's prior.
Finally, this modified scene graph is given as input to MetaSim, and we train MetaSim
to transform the modified attributes back to their original ones.
In our setting, this training process was both faster and more stable than the
original unsupervised algorithm.

\begin{table*}[!t]
\centering
\small
\begin{tabular}{l c c c c c c}
\toprule
              & \multicolumn{3}{c}{{\ourdataset}}               & \multicolumn{3}{c}{Argoverse} \\
              \cmidrule(lr){2-4}                                \cmidrule(lr){4-7}
Method        & Size          & Speed         & Heading         & Size          & Speed         & Heading \\
\midrule
Prob. Grammar & 0.49          & 0.42          & 0.30            & 0.41          & 0.57          & 0.38 \\
MetaSim       & 0.49          & 0.33          & 0.14            & 0.50          & 0.53          & \textbf{0.18} \\
Procedural    & 0.15          & 0.41          & \textbf{0.07}   & 0.23          & 0.59          & \textbf{0.17} \\
Lane Graph    & 0.33          & 0.28          & 0.16            & 0.31          & 0.34          & 0.38 \\
LayoutVAE     & 0.16          & 0.40          & 0.29            & 0.21          & 0.46          & 0.29 \\
\midrule
SceneGen      & \textbf{0.06} & \textbf{0.19} & 0.08            & \textbf{0.15} & \textbf{0.20} & 0.22\\
\bottomrule
\end{tabular}
\caption{
Vehicle-only maximum mean discrepency (MMD) results
on {\ourdataset} and Argoverse.
}
\label{table:quantitative-vehicle-only}
\end{table*}

\paragraph{Procedural:}
Our Procedural baseline is inspired by methods that operate directly on the
road topology of the traffic scene~\cite{wheeler2015,wheeler2016,jesenski2019,manivasagam2020}.
Specifically, given a \emph{lane graph} of the scene~\cite{liang2020},
Procedural uses a set of rules to place actors onto lane centerlines.
First, we determine a set of valid routes traversing the entire
lane graph.
Each valid route is a sequence of successive lane centerlines along which actors
can traverse without violating traffic rules; \eg, running red lights, merging
onto an oncoming lane, \etc.
Next, we place actors onto each route such that successive actors maintain
a random clearance (drawn from an exponential distribution) and no two actors
collide.
Each actor's bounding box size is sampled form a Gaussian KDE fitted to the
training dataset, and its heading is determined by the tangent vector along
its lane centerline at its location.
Finally, we initialize the speed of each actor such that successive actors
maintain a random time gap (drawn from an exponential distribution).
Procedural is similar to the heuristics underlying \cite{wheeler2015,wheeler2016,jesenski2019}
but generalized to handle arbitrary road topologies.
Similar to Prob. Grammar, Procedural can generate only vehicles and
bicyclists since existing HD maps do not provide sidewalks.
We believe this limitation highlights the difficulty of using a heuristics-based approach.

\paragraph{Lane Graph:}
Inspired by MetaSim, we also consider a learning-based version of Procedural.
Specifically, given a traffic scene generated by Procedural, we use a lane
graph neural network to transform the attributes of each actor; \ie, location,
bounding box size, heading, and velocity.
Our lane graph neural network follows the design of the state-of-the-art
motion forecasting model by Liang~\etal~\cite{liang2020}.
It consists of MapNet for extracting map topology features and four fusion
modules: actor-to-lane, lane-to-lane, lane-to-actor, and actor-to-actor.
We train Lane Graph using heuristically generated ground truth, as in our
MetaSim baseline.

\paragraph{LayoutVAE:}
Our implementation of LayoutVAE largely follows that of Jyothi \etal~\cite{jyothi2019}.
To adapt LayoutVAE to traffic scene generation, we first augment the original model
with an additional CNN to extract map features.
In particular, given a bird's eye view multi-channel image of the HD map, we use
the backbone architecture of Liang~\etal~\cite{liang2020a} to extract multi-scale
map features, which we subsequently average-pool into a feature vector.
This is then given to LayoutVAE as input in place of the label set encoding
used in the original setting\footnote{The label set in our setting is fixed
to be vehicles, pedestrians, and bicyclists.}.
Our second modification enables LayoutVAE to output oriented bounding boxes
and velocities.
Specifically, we replace the spherical quadrivariate Gaussian distribution of
its BBoxVAE with a bivariate Gaussian distribution for location, a bivariate
log-normal distribution for bounding box size, and a bivariate Gaussian
distribution for velocity.
To evaluate the log-likelihood of a scene, we use Monte-Carlo approximation
with 1000 samples from the conditional prior~\cite{jyothi2019}.

\subsection{MMD Metrics}
To complement our likelihood-based metric, we compute a sample-based metric as
well: maximum mean discrepancy (MMD)~\cite{gretton2012}.
As we discussed in the main text, MMD measures a distance between two
distributions $ p $ and $ q $ as
\begin{align}
    \mathrm{MMD}^2(p, q) = \mathbb{E}_{x, x' \sim p}[k(x, x')]
    + \mathbb{E}_{y, y' \sim q}[k(y, y')] - 2\mathbb{E}_{x \sim p, y \sim q}[k(x, y)]
\end{align}
for some kernel $ k $.
Following~\cite{you2018,liao2019}, we compute MMD using Gaussian kernels
(with bandwidth $ \sigma = 1 $) with the total variation distance to compare
scene statistics between generated and real traffic scenes.
In particular, we first sample a set $ P $ of real traffic scenes from
the evaluation dataset.
Conditioned the SDV state and HD map of the scenes in $ P $, we generate
a set $ Q $ of synthetic scenes using the method under evaluation.
Then, we approximate MMD as:
\begin{align}
    \mathrm{MMD}^2(p, q) \approx
    \frac{1}{|P|^2}\sum_{x \in P} \sum_{x' \in P} k(x, x')
    + \frac{1}{|Q|^2} \sum_{y \in Q} \sum_{y' \in Q} k(y, y')
    - \frac{2}{|P||Q|} \sum_{x \in P} \sum_{y \in Q} k(x, y)
\end{align}

Our scene statistics measure the distribution of classes,
bounding box sizes (in $ m^2 $), speeds (in $ m / s $),
and heading angles (relative to that of the SDV) for each scene.
Empty scenes are discarded since these scene statistics are undefined.
Since MMD is expensive to compute, in {\ourdataset}, we form $ P $ by
sampling the evaluation dataset by every 25th scene, yielding approximately 5000 scenes.
We compute MMD over the full Argoverse validation set as it contains
5015 scenes only.

We also compute MMD in the feature space of a pre-trained motion forecasting model.
This is similar to some popular metrics for evaluating generative models such
IS~\cite{salimans2016}, FID~\cite{heusel2017}, and KID~\cite{binkowski2018},
except we use a motion forecasting model as our feature extractor.
Here, our motion forecasting model takes a bird's eye view multi-channel image
of the actors in the scene and regresses the future locations of each actor
over the next 3 seconds in 0.5s increments.
We use the actor rasterization procedure described in Sec.~\ref{section:method/input-representation}
and the model architecture from \cite{wong2020}, and we train the model using
4000 training log from the {\ourdataset} training set.
To obtain a feature vector summarizing the scene, we average pool the
model's backbone features along its spatial dimensions.
Then, to compute MMD, we use the RBF kernel with bandwidth $ \sigma = 1 $.

\section{Additional Experiment Results}
\label{section:quantiative}

\begin{figure}
\centering
\includegraphics[width=0.75\textwidth]{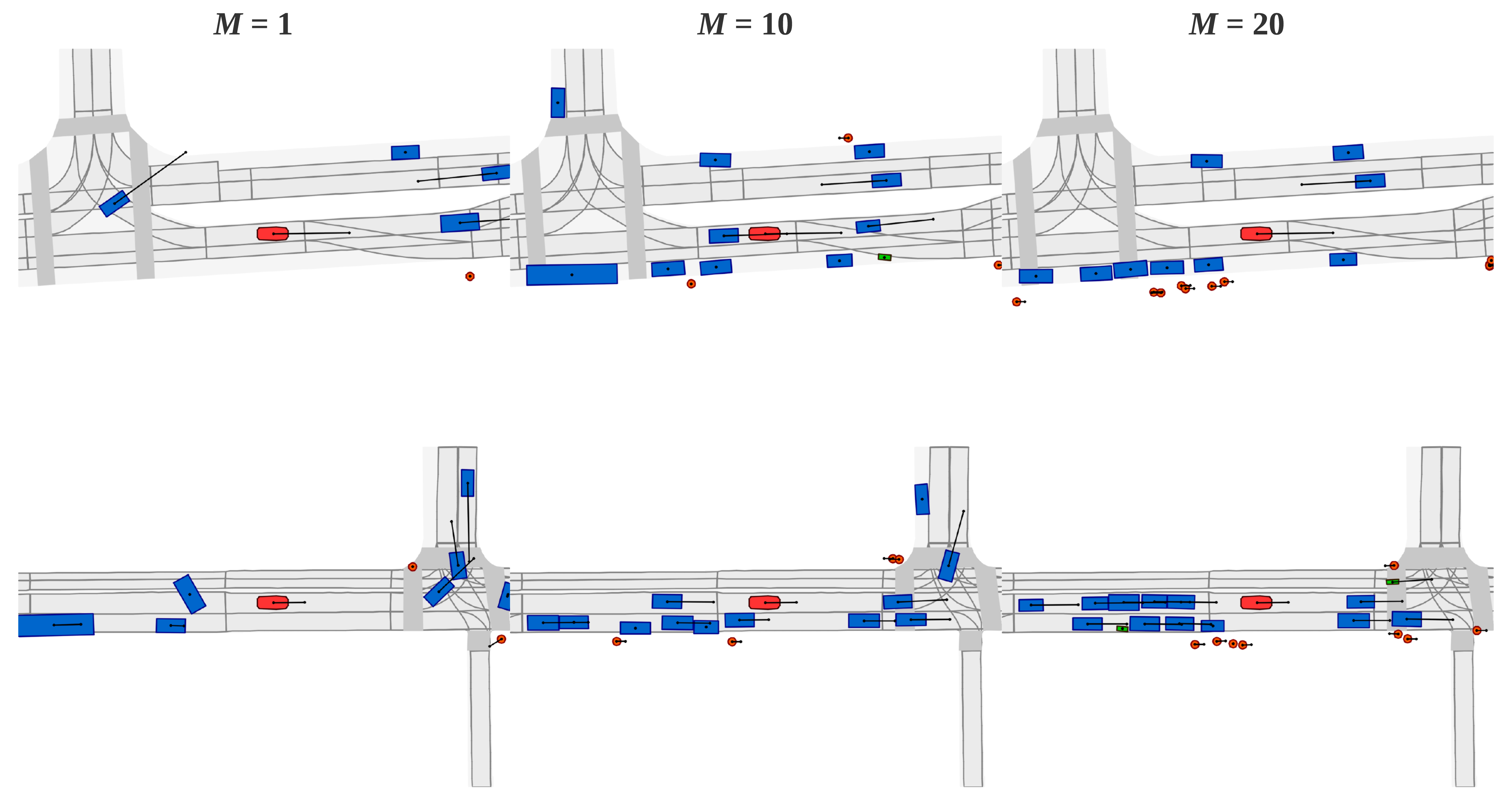}
\caption{Traffic scenes generated by {\modelname} using $ M = 1, 10, 20 $ sample proposals for {\ourdataset}.}
\label{figure:sampling-qualitative-results}
\end{figure}

\begin{table*}[!t]
\centering
\small
\begin{tabular}{l c c c c c c c}
\toprule
$ M $         & Features & Class & Size & Speed & Heading \\
\midrule
1             & 0.13     & 0.05  & 0.05 & 0.10  & 0.10    \\
10            & 0.11     & 0.20  & 0.06 & 0.33  & 0.08    \\
20            & 0.11     & 0.30  & 0.07 & 0.41  & 0.08    \\
\bottomrule
\end{tabular}
\caption{
Analysis of the number of sample proposals $ M $ on {\ourdataset}.
The reported numbers are the MMD metrics computed between distributions
of features extracted by a motion forecasting model and various scene
statistics (see main text).
}
\label{table:quantitative-sampling-ablation}
\end{table*}

\subsection{Vehicle MMD Metrics}
In Tab.~\ref{table:quantitative-vehicle-only}, we report vehicle-only MMD
metrics for {\ourdataset} and Argoverse.
Specifically, we compute scene statistics for generated and real traffic
scenes using vehicle actors only.
As before, scenes with no vehicles are discarded during evaluation.
This allows for an alternative comparison that controls
for the class most easily handled by heuristics; \ie, vehicles.
Overall, we see that {\modelname} still achieves the best results among the
competing methods.
This result reaffirms our claim that heuristics-based methods are insufficient
to model the full complexity and diversity of real world traffic scenes.

\subsection{Sampling Strategy Analysis}
As discussed in the main text, {\modelname} uses a sampling strategy inspired by
\emph{nucleus sampling}~\cite{holtzman2020}.
Specifically, at each generation step, we sample each of SceneGen's
position, heading, and velocity distributions $ M $ times and return
the most likely sample as output.
In Tab.~\ref{table:quantitative-sampling-ablation} and Fig.~\ref{figure:sampling-qualitative-results},
we analyze the effects of using different numbers of sample proposals $ M = 1, 10, 20 $.
We see that using $ M > 1 $ decreases MMD on deep features, indicating that
scene-level realism is improved.
This improvement is even more evident in Fig.~\ref{figure:sampling-qualitative-results},
where we see vehicles disregarding the rules of traffic when $ M = 1 $.
With more fine-grained tuning of $ M $, we expect to see improvements in
the actor-level statistics as well; \ie, class, size, and speed.

\section{Additional Qualitative Results}
\label{section:qualitative}
\begin{figure*}
\includegraphics[width=\textwidth]{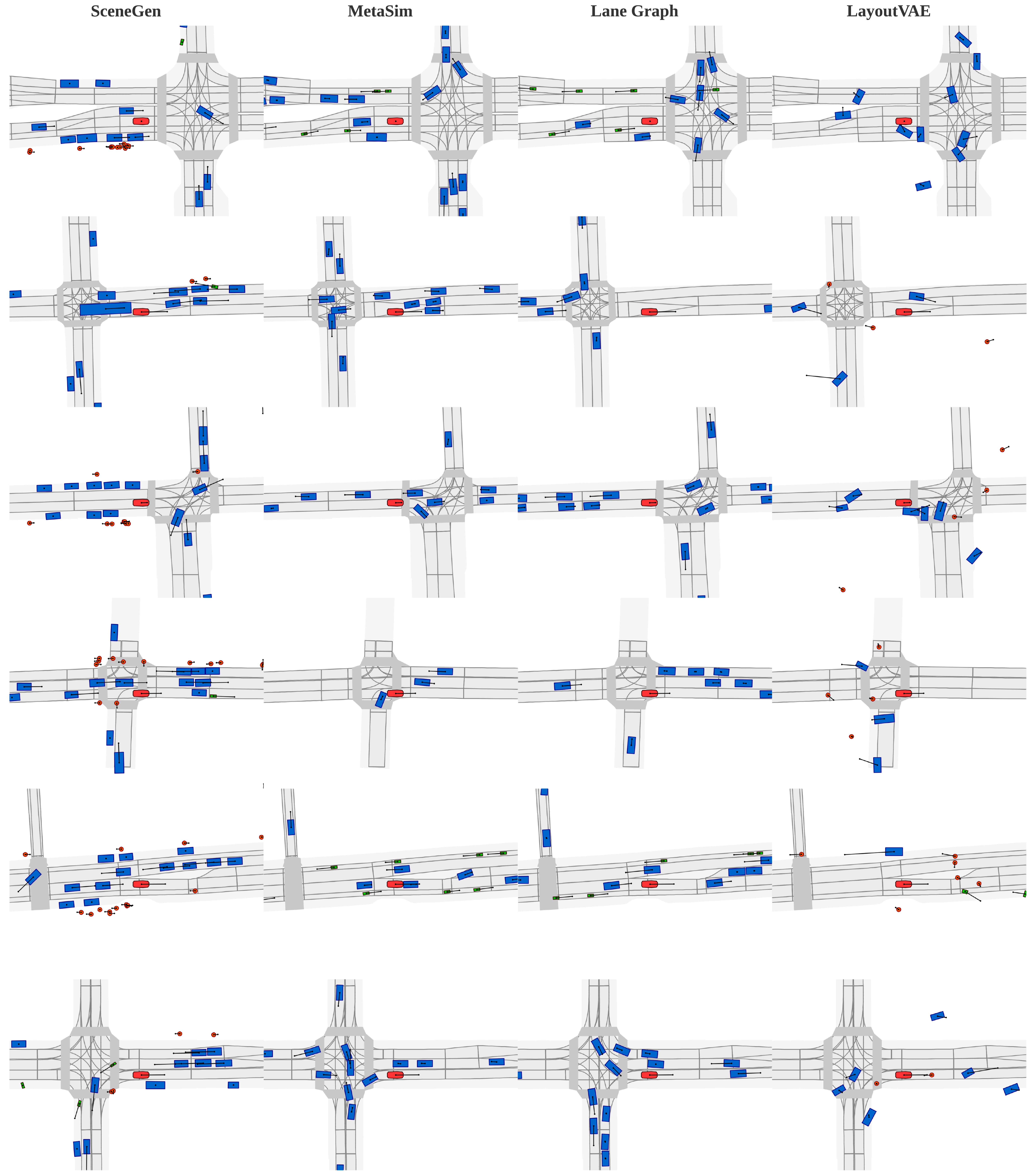}
\caption{
Qualitative comparison of traffic scenes generated by SceneGen and various
baselines on {\ourdataset}.
The ego SDV is shown in \textcolor{red}{red};
vehicles in \textcolor{blue}{blue};
pedestrians in \textcolor{orange}{orange};
and bicyclists in \textcolor{green}{green}.
We visualize lane segments and drivable surfaces in light grey
and crosswalks in dark grey.
}
\label{figure:atg4d-qualitative-results}
\end{figure*}

\begin{figure*}
\includegraphics[width=\textwidth]{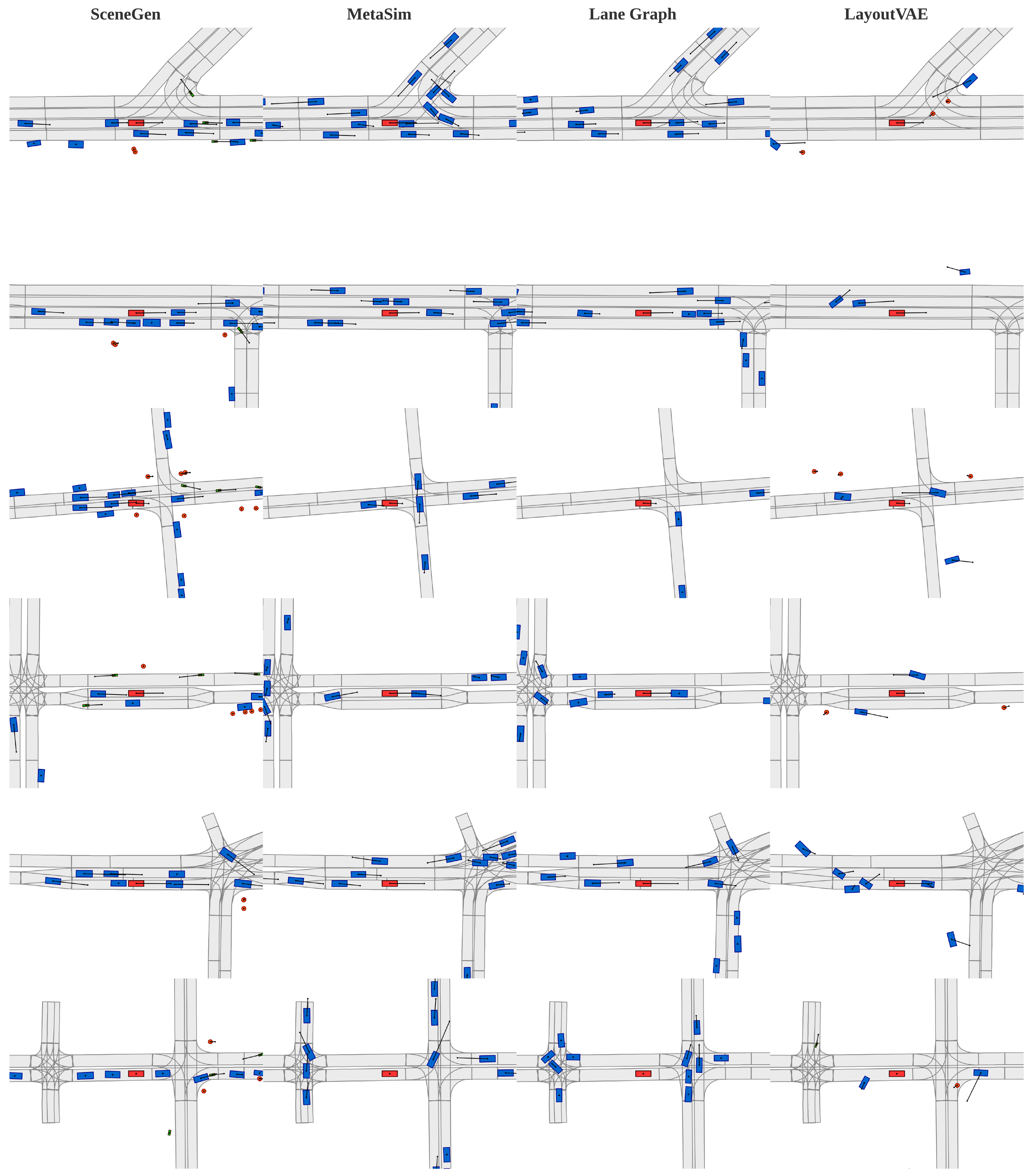}
\caption{
Qualitative comparison of traffic scenes generated by SceneGen and various
baselines on Argoverse.
The ego SDV is shown in \textcolor{red}{red};
vehicles in \textcolor{blue}{blue};
pedestrians in \textcolor{orange}{orange};
and bicyclists in \textcolor{green}{green}.
We visualize lane segments in light grey.
}
\label{figure:argoverse-qualitative-results}
\vspace{10pt}
\end{figure*}

In Fig.~\ref{figure:atg4d-qualitative-results} and \ref{figure:argoverse-qualitative-results},
we present an array of additional qualitative results for {\ourdataset} and Argoverse respectively.
Here, we compare traffic scenes generated by {\modelname}, MetaSim,
Lane Graph, and LayoutVAE.
From these visualizations, we see that {\modelname} generates traffic
scenes that best reflect the complexity and diversity of real world
traffic scenes.
For example, in the second-to-last row of Fig.~\ref{figure:atg4d-qualitative-results},
we show a traffic scene generated by {\modelname} in which a vehicle
performs a three-point turn.
In the bottom row of Fig.~\ref{figure:atg4d-qualitative-results},
we also show a scene in which two bicyclists
perform an left turn using the car lane.
These scenes highlight {\modelname}'s ability to model rare
but plausible traffic scenes that could occur in the real world.

\begin{figure*}
\includegraphics[width=\textwidth]{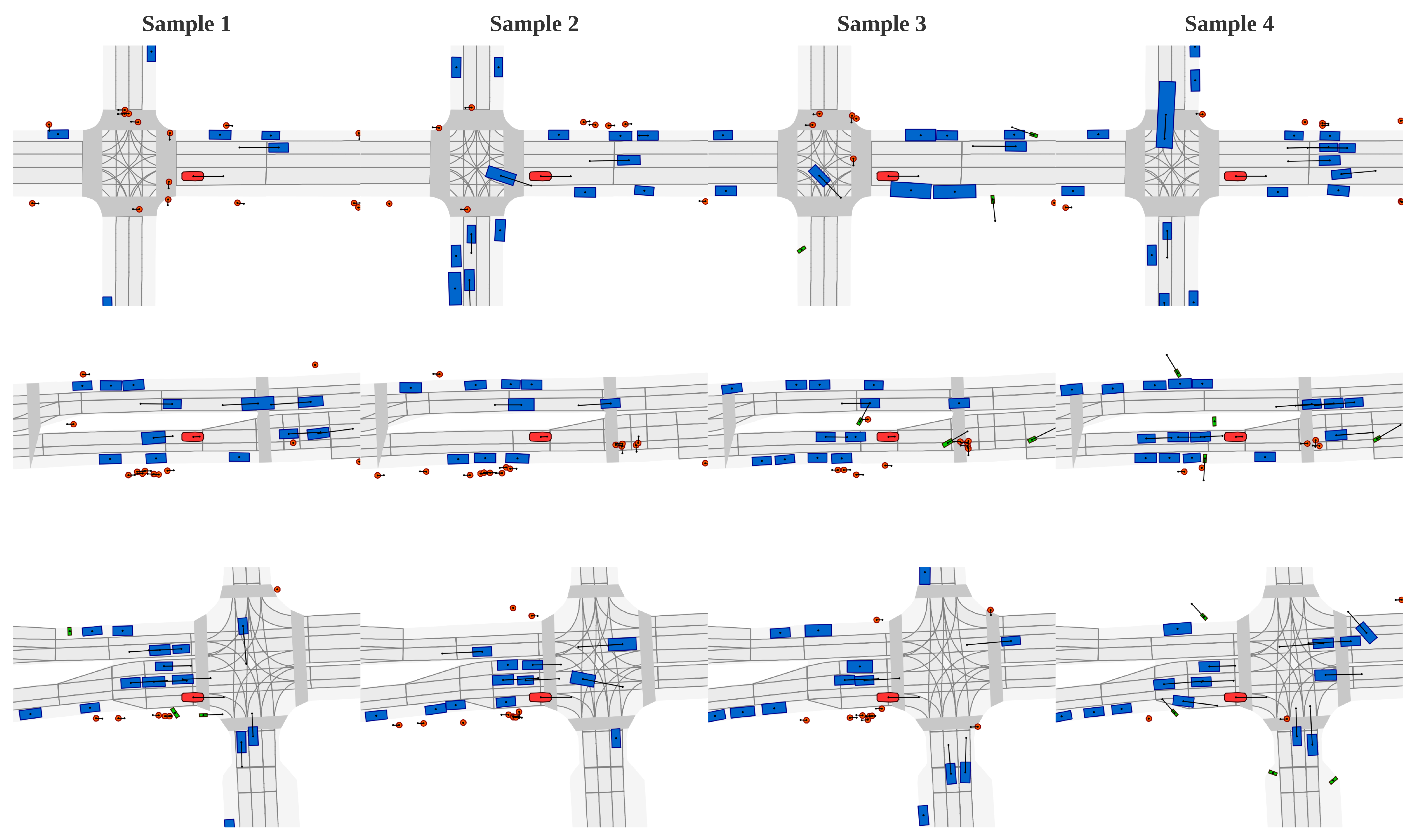}
\caption{
Traffic scenes generated by {\modelname} on {\ourdataset}.
The traffic scenes in each row are generated from the same SDV state and HD map inputs.
Each traffic scene is a distinct sample drawn from our model.
}
\label{figure:atg4d-diversity-results}

\includegraphics[width=\textwidth]{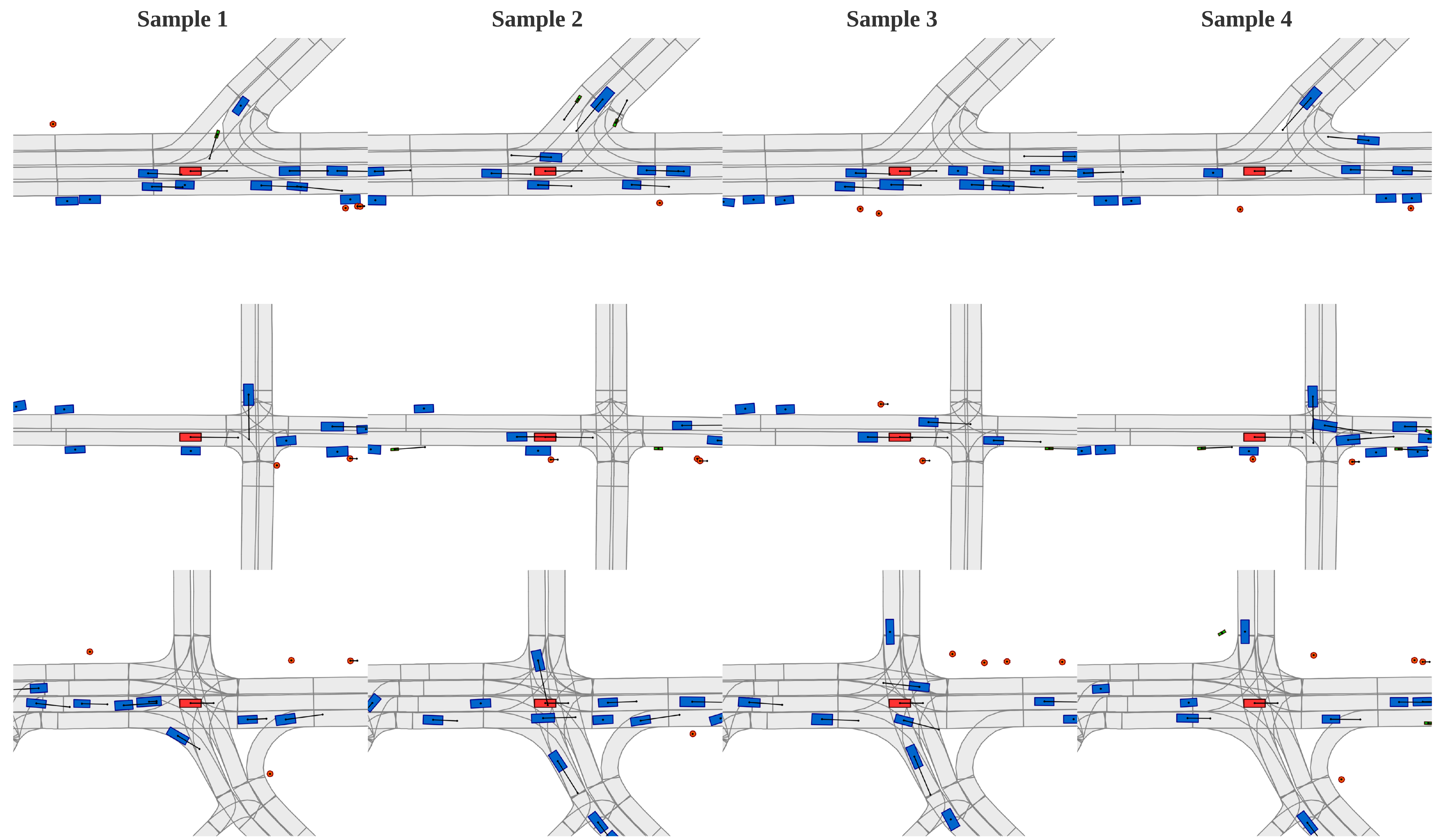}
\caption{
Traffic scenes generated by {\modelname} on Argoverse.
The traffic scenes in each row are generated from the same SDV state and HD map inputs.
Each traffic scene is a distinct sample drawn from our model.
}
\label{figure:argoverse-diversity-results}
\end{figure*}

In Fig.~\ref{figure:atg4d-diversity-results} and \ref{figure:argoverse-diversity-results},
we also showcase the diversity of traffic scenes that {\modelname} is able to generate.
Each row in the figures show four samples from our model when given the same
SDV state and HD map as inputs.
From these visualizations, we see that {\modelname} captures the multi-modality
of real world traffic scenes well.
For example, the top row of Fig.~\ref{figure:atg4d-diversity-results} shows
four traffic scenes generated for a four-way intersection.
Here, we see samples in which pedestrians cross the intersection, vehicles
perform an unprotected left turn, and a large bus goes straight.

\begin{figure*}
\includegraphics[width=\textwidth]{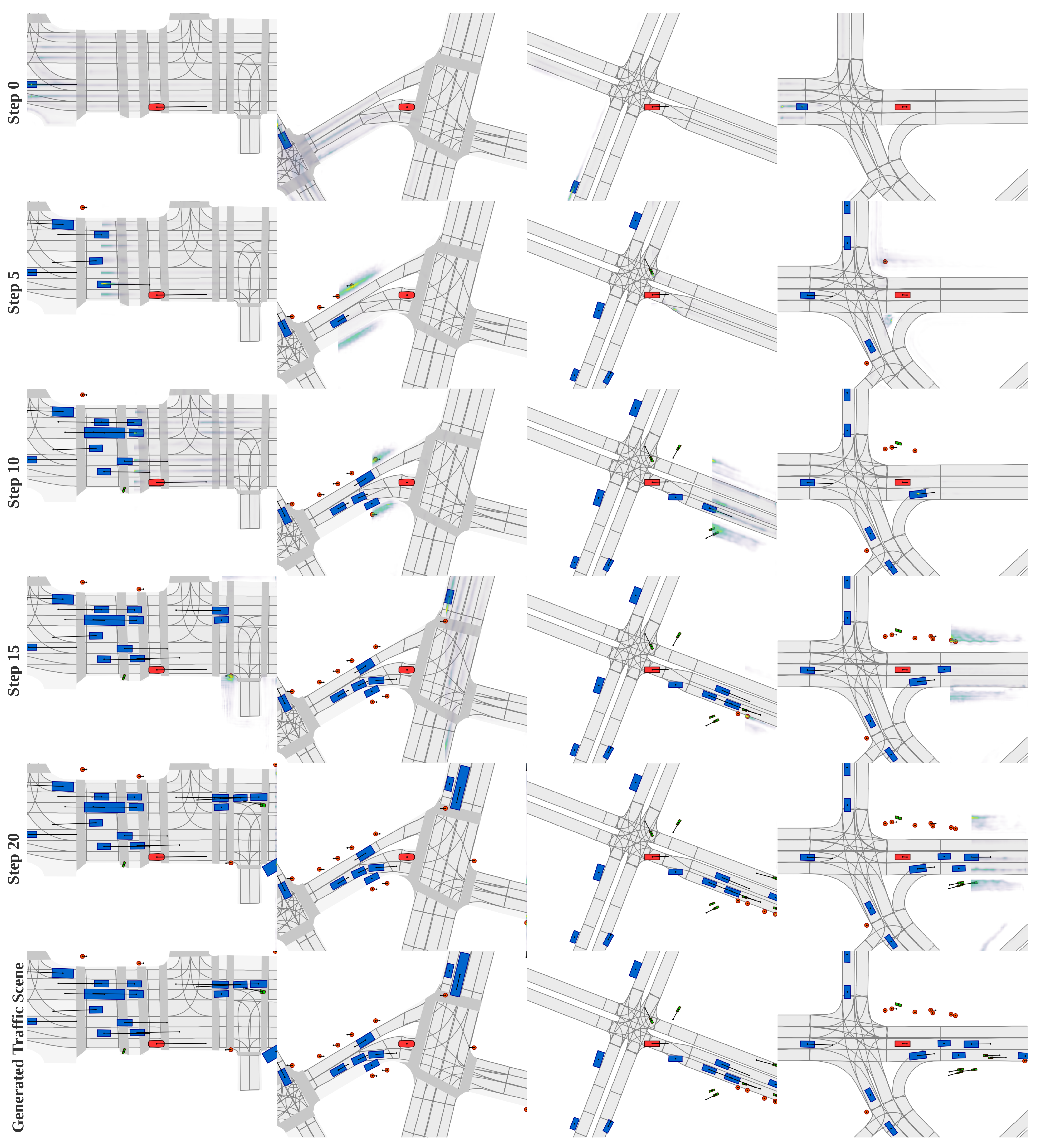}
\caption{
Traffic scenes generated by {\modelname} on {\ourdataset} (first two columns)
and Argoverse (last two columns).
We visualize the quantized location heatmap for steps $ t = 0, 5, 10, 15, 20 $
of the generation process.
Each column represents the generation process for one traffic scene.
Bright yellow means higher likelihood.
}
\label{figure:heatmap-results}
\vspace{10pt}
\end{figure*}

Finally, in Fig.~\ref{figure:heatmap-results}, we visualize the quantized
location heatmaps for steps $ t = 0, 5, 10, 15, 20 $ of the generation process.
Each row shows the categorical distribution from which we sample the next
actor's location.
From these visualizations, we see that {\modelname} is able to model the
distribution over actor locations (and the corresponding uncertainties) quite
precisely.
For example, the distribution over vehicle locations are concentrated around
lane centerlines and the distribution over pedestrian locations are diffused
over crosswalks and sidewalks.

\newpage
{\small
\bibliographystyle{ieee_fullname}
\bibliography{egbib}
}